\documentclass{article}

\usepackage{arxiv}

\usepackage[utf8]{inputenc}
\usepackage[T1]{fontenc}
\usepackage[
    backend=biber,
    style=numeric-comp,
    sorting=nty,
    natbib=true,
    mincitenames=1,
    maxcitenames=1,
    maxbibnames=3
]{biblatex}
\usepackage{xurl}
\usepackage{booktabs}
\usepackage{amsmath}
\usepackage{amsfonts}
\usepackage{nicefrac}
\usepackage{microtype}
\usepackage{xcolor}
\definecolor{citationblue}{rgb}{0.09,0.5,0.99}
\usepackage{hyperref}
\usepackage{colortbl}
\usepackage{makecell}
\usepackage{enumitem}
\usepackage{subcaption}
\usepackage{mathrsfs}
\usepackage{multirow}
\usepackage{algorithm}
\usepackage{algorithmic}
\usepackage[nameinlink, capitalise]{cleveref}
\crefname{figure}{Figure}{Figures}
\Crefname{figure}{Figure}{Figures}
\crefname{table}{Table}{Tables}
\Crefname{table}{Table}{Tables}
\usepackage{wrapfig}
\usepackage{graphicx}

\newcommand{\cypatablefont}{\footnotesize}
\newif\ificlrversion
\iclrversionfalse
\newenvironment{cypawidetable}{}{}

\title{CyFA: Linear Sequence Modeling\\with Relative-Time-Partitioned Memory}
\author{%
    Yixiao Chen\\
    {\normalfont\small Tsinghua University}\\
    {\normalfont\footnotesize\texttt{chenyixi22@mails.tsinghua.edu.cn}}
    \And
    Shuojin Yang\\
    {\normalfont\small Tsinghua University}\\
    {\normalfont\footnotesize\texttt{yangshuojin@mail.com.cn}}
    \And
    Shi-Min Hu\\
    {\normalfont\small Tsinghua University}\\
    {\normalfont\footnotesize\texttt{shimin@tsinghua.edu.cn}}
}
\date{}
\renewcommand{\shorttitle}{CyFA}
\renewcommand{\headeright}{}
\renewcommand{\undertitle}{}

\hypersetup{
    colorlinks=true,
    linkcolor=black,
    citecolor=citationblue,
    urlcolor=black,
    pdftitle={CyFA: Linear Sequence Modeling with Relative-Time-Partitioned Memory},
    pdfauthor={Yixiao Chen, Shuojin Yang, Shi-Min Hu}
}

\begin{document}

\maketitle
\pagestyle{fancy}

\begin{abstract}
Linear RNNs offer linear-time sequence processing and constant-memory decoding,
but their fixed-size recurrent states must accommodate all past key--value
associations. Existing forgetting mechanisms and Delta Rule updates reduce
interference by selectively clearing or correcting the state, yet earlier
associations can still become difficult to retrieve. We introduce CyFA (Cyclic
Flow Attention), a Linear RNN with relative-time-partitioned memory. At each
step, a learned clock controls the cyclic transport applied jointly to the key
and value states before the current key--value pair enters the age-zero slot,
thereby organizing stored associations across relative-time slots. We further
derive an exact change to absolute-clock coordinates that expresses CyFA as two
scalar-decay linear attention recurrences and enables efficient chunk-wise
training. Across 400M--1.4B pretraining experiments with matched
recurrent-state sizes, CyFA improves recall-intensive performance while
maintaining competitive language modeling and high computational efficiency.
At 400M, CyFA outperforms KDA on FDA ($42.60$ vs. $26.07$) while requiring only
$46.7\%$ and $48.3\%$ of KDA's forward and backward core-operator execution
times, respectively. Our code is publicly
available at \url{https://github.com/Chyxx/CyclicFlowAttention}.
\end{abstract}

\section{Introduction}

Transformers have become the dominant architecture for sequence modeling by
combining scalable parallel training with expressive content-dependent
interactions \citep{vaswani2017attention}. At long sequence lengths, however,
full attention incurs quadratic computation, while autoregressive decoding
maintains a key--value cache that grows linearly with context length. Linear
RNNs, including linear attention models and modern state space models, provide
an efficient alternative by compressing the sequence prefix into a fixed-size
recurrent state. This enables linear-time sequence processing and
constant-memory decoding
\citep{katharopoulos2020transformers,gu2022s4,dao2024mamba2}. The same
efficiency imposes a fixed-state constraint: every new write updates the same
recurrent state. In additive linear attention, an increasing number of
key--value associations therefore share this state and interfere with one
another \citep{schlag2021fastweight}.

Modern Linear RNNs mitigate this interference by learning what to erase from
recurrent memory as new writes arrive. Forget gates attenuate the existing
state, while Delta Rule updates erase only the association addressed by the
incoming key before writing the new value
\citep{gu2023mamba,yang2024gla,schlag2021fastweight,
yang2024deltanetparallel}. These mechanisms have substantially improved Linear
RNNs, but recall remains difficult when many past associations must remain
simultaneously retrievable.

A natural response is to make memory erasure more selective. Recent models use
increasingly fine-grained forget gates and more expressive Delta Rule updates,
reducing unnecessary information loss but adding computation to the recurrence
and requiring more involved chunk-wise algorithms
\citep{kimiteam2025kimi,hatamizadeh2026gdn2,peng2025rwkv7}. Both directions
continue to focus on deciding what existing information to erase when a new
write arrives. This raises a complementary question: \textbf{can the memory
itself be better organized so that, under the same fixed state budget, more
associations remain distinguishable and retrievable before they must be
forgotten?}

\begin{figure}[t]
    \centering
    \includegraphics[width=\linewidth]{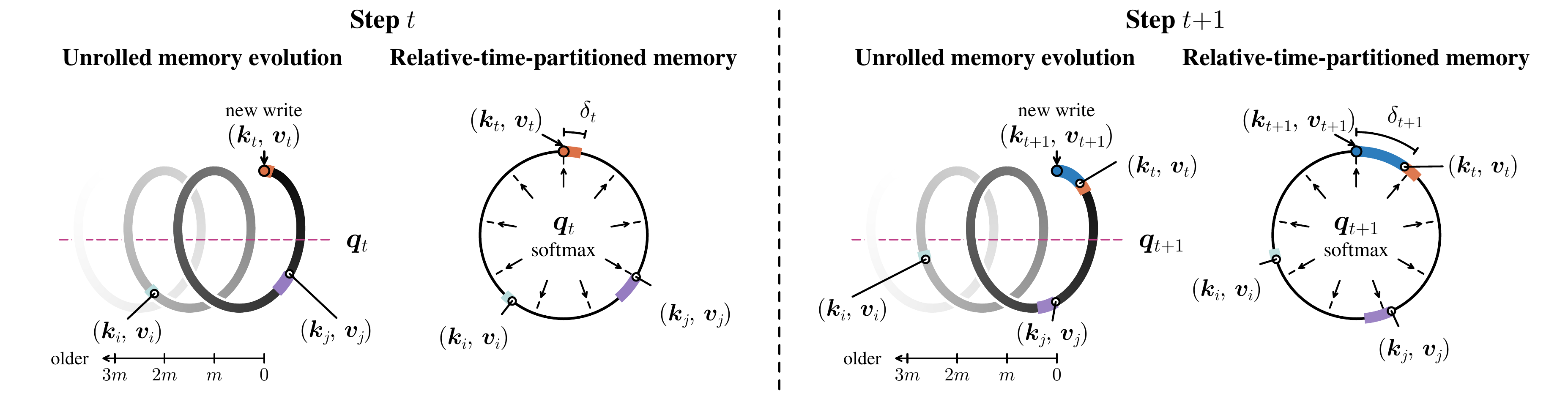}
    \caption{\textbf{Cyclic transport in CyFA.}
    The left panels unroll cyclic transport across successive cycles to
    visualize how stored associations advance with model age as new writes
    enter at age zero. In the recurrent state, successive cycles reuse the same
    finite set of relative-time slots, causing their contributions to overlap
    as shown in the right panels. The learned clock increment $\delta_t$
    controls the transport at each step, and the query $\boldsymbol q_t$
    produces softmax weights over the resulting slots. Lighter colors indicate
    scalar decay. We omit the readout matrix $\mathbf R$ and write strength
    $\beta_t$ for clarity.}
    \label{fig:cyfa-helical-memory}
\end{figure}

We introduce Cyclic Flow Attention (CyFA), a Linear RNN with
relative-time-partitioned memory. Before every new write, CyFA applies the same
cyclic transport to its key and value states, preserving their alignment as
stored associations advance through the relative-time slots. The current
key--value pair is then written to the age-zero slot. The transport accumulated
since a write entered memory defines its model age. Recent writes remain near
age zero, while earlier writes advance around the cycle. A learned clock makes
this transport input dependent by controlling the amount applied at each token,
thereby adapting the spacing between successive writes.
Figure~\ref{fig:cyfa-helical-memory} illustrates this memory evolution.

% Introduction figure. Paths are relative to the paper's main .tex file.

Directly applying cyclic transport would mix all relative-time slots at every
token, creating a dense sequential transition. To avoid this cost, we derive an
exact change to absolute-clock coordinates that expresses the same update as
two scalar-decay recurrences connected by a token-wise readout. This formulation
enables hardware-efficient chunk-wise training. Across 400M--1.4B models with
matched main recurrent state sizes, CyFA maintains competitive language
modeling and achieves $5.74$--$6.94$ percentage points higher average accuracy
on recall-intensive tasks than KDA (\cref{tab:language_recall_results}). Using
the 400M training shapes, CyFA pairs these gains with $1.37\times$ KDA's
end-to-end training throughput, while requiring only $46.7\%$ and $48.3\%$ of
KDA's forward and backward core-operator execution times, respectively
(\cref{fig:efficiency_400m,fig:kernel_latency_400m}).

\section{Background and Preliminaries}
\label{sec:background}

\subsection{Linear Attention}
\label{sec:background-linear-attention}

Linear attention compresses the key--value prefix into a fixed-size associative state. Let $\boldsymbol q_t,\boldsymbol k_t\in\mathbb R^{d_k}$, $\boldsymbol v_t\in\mathbb R^{d_v}$, and $\mathbf S_t\in\mathbb R^{d_k\times d_v}$. Its causal recurrence and readout are
\begin{equation}
\begin{aligned}
    \mathbf S_t
    &= \mathbf S_{t-1}+\boldsymbol k_t\boldsymbol v_t^\top,\\
    \boldsymbol o_t
    &= \mathbf S_t^\top\boldsymbol q_t
     = \sum_{i\le t}\boldsymbol v_i(\boldsymbol k_i^\top\boldsymbol q_t).
\end{aligned}
\label{eq:background-linear-attention}
\end{equation}
The unrolled readout shows that all writes up to position $t$ are retrieved through the same state \citep{katharopoulos2020transformers}.

As writes accumulate, associations stored in the same state can interfere \citep{schlag2021fastweight}. Forgetting mechanisms mitigate this interference by erasing part of the existing memory before each new write. Mamba-2 uses a data-dependent scalar forget gate $\alpha_t$, which uniformly decays the previous state \citep{dao2024mamba2}. With write strength $\beta_t$, the recurrence becomes
\begin{equation}
    \mathbf S_t
    = \alpha_t\mathbf S_{t-1}
    + \beta_t\boldsymbol k_t\boldsymbol v_t^\top.
\label{eq:background-scalar-forgetting}
\end{equation}
The Delta Rule instead makes erasure key dependent \citep{schlag2021fastweight,yang2024deltanetparallel}. The incoming key identifies the stored association to be erased, and $\beta_t$ controls how strongly that association is moved toward the new value:
\begin{equation}
    \mathbf S_t
    = \mathbf S_{t-1}
    + \beta_t\boldsymbol k_t
      \bigl(\boldsymbol v_t-\mathbf S_{t-1}^\top\boldsymbol k_t\bigr)^\top.
\label{eq:background-delta-rule}
\end{equation}
Gated DeltaNet (GDN) and Kimi Delta Attention (KDA) combine this key-dependent erasure with scalar and vector-valued forget gates, respectively \citep{yang2024gateddelta,kimiteam2025kimi}.

\begin{table}[!t]
\centering
\caption{Representative linear attention recurrences and interfaces together with their relative training costs. Write gates for RetNet and Mamba-2 are omitted because they can be absorbed into $\boldsymbol k_t$ or $\boldsymbol v_t$. GLA left-multiplies the state with row-wise decay, whereas GLA (col) denotes the corresponding column-wise form that right-multiplies the state.}
\label{tab:memory-recurrences}
\cypatablefont
\setlength{\tabcolsep}{2pt}
\renewcommand{\arraystretch}{1.28}
\begin{cypawidetable}
\begin{tabular}{lllll}
\toprule
\textbf{Method} & \textbf{State Update} & \textbf{Readout} & \textbf{Linear Attention Interface} & \textbf{Cost} \\
\midrule
LA
& $\mathbf S_t=\mathbf S_{t-1}+\boldsymbol k_t\boldsymbol v_t^\top$
& $\boldsymbol o_t=\mathbf S_t^\top\boldsymbol q_t$
& $\operatorname{LA}(\{\boldsymbol q_t,\boldsymbol k_t,\boldsymbol v_t\}_{t=1}^{T})$ & Low \\
RetNet
& $\mathbf S_t=\gamma\mathbf S_{t-1}+\boldsymbol k_t\boldsymbol v_t^\top$
& $\boldsymbol o_t=\mathbf S_t^\top\boldsymbol q_t$
& $\operatorname{FixedScalarGatedLA}(\{\boldsymbol q_t,\boldsymbol k_t,\boldsymbol v_t,\gamma\}_{t=1}^{T})$ & Low \\
Mamba-2
& $\mathbf S_t=\alpha_t\mathbf S_{t-1}+\boldsymbol k_t\boldsymbol v_t^\top$
& $\boldsymbol o_t=\mathbf S_t^\top\boldsymbol q_t$
& $\operatorname{ScalarGatedLA}(\{\boldsymbol q_t,\boldsymbol k_t,\boldsymbol v_t,\alpha_t\}_{t=1}^{T})$ & Low \\
GLA
& $\mathbf S_t=\operatorname{Diag}(\boldsymbol\alpha_t)\mathbf S_{t-1}+\boldsymbol k_t\boldsymbol v_t^\top$
& $\boldsymbol o_t=\mathbf S_t^\top\boldsymbol q_t$
& $\operatorname{RowGatedLA}(\{\boldsymbol q_t,\boldsymbol k_t,\boldsymbol v_t,\boldsymbol\alpha_t\}_{t=1}^{T})$ & Mid \\
GLA (col)
& $\mathbf S_t=\mathbf S_{t-1}\operatorname{Diag}(\boldsymbol\alpha_t)+\boldsymbol k_t\boldsymbol v_t^\top$
& $\boldsymbol o_t=\mathbf S_t^\top\boldsymbol q_t$
& $\operatorname{ColumnGatedLA}(\{\boldsymbol q_t,\boldsymbol k_t,\boldsymbol v_t,\boldsymbol\alpha_t\}_{t=1}^{T})$ & Mid \\
DeltaNet
& $\mathbf S_t=(\mathbf I-\beta_t\boldsymbol k_t\boldsymbol k_t^\top)\mathbf S_{t-1}+\beta_t\boldsymbol k_t\boldsymbol v_t^\top$
& $\boldsymbol o_t=\mathbf S_t^\top\boldsymbol q_t$
& $\operatorname{DeltaRule}(\{\boldsymbol q_t,\boldsymbol k_t,\boldsymbol v_t,\beta_t\}_{t=1}^{T})$ & Mid \\
GDN
& $\mathbf S_t=\alpha_t(\mathbf I-\beta_t\boldsymbol k_t\boldsymbol k_t^\top)\mathbf S_{t-1}+\beta_t\boldsymbol k_t\boldsymbol v_t^\top$
& $\boldsymbol o_t=\mathbf S_t^\top\boldsymbol q_t$
& $\operatorname{ScalarGatedDeltaRule}(\{\boldsymbol q_t,\boldsymbol k_t,\boldsymbol v_t,\beta_t,\alpha_t\}_{t=1}^{T})$ & Mid \\
KDA
& $\mathbf S_t=(\mathbf I-\beta_t\boldsymbol k_t\boldsymbol k_t^\top)\operatorname{Diag}(\boldsymbol\alpha_t)\mathbf S_{t-1}+\beta_t\boldsymbol k_t\boldsymbol v_t^\top$
& $\boldsymbol o_t=\mathbf S_t^\top\boldsymbol q_t$
& $\operatorname{RowGatedDeltaRule}(\{\boldsymbol q_t,\boldsymbol k_t,\boldsymbol v_t,\beta_t,\boldsymbol\alpha_t\}_{t=1}^{T})$ & High \\
\bottomrule
\end{tabular}
\end{cypawidetable}
\end{table}

\paragraph{Chunk-wise parallelism.}
These token-wise recurrences support efficient autoregressive decoding, but their sequential dependencies limit training parallelism. Chunk-wise algorithms address this limitation by propagating the recurrent state only across chunk boundaries while computing the outputs within each chunk in parallel \citep{yang2024fla}. Let each chunk contain $C$ tokens, let $\mathbf S_{[i]}=\mathbf S_{iC}$ denote the state after chunk $i$, and let $\mathbf Q_{[i+1]},\mathbf K_{[i+1]},\mathbf V_{[i+1]}$ collect the queries, keys, and values in the next chunk. For additive linear attention, with within-chunk causal mask $\mathbf M$, the boundary state and outputs are
\begin{equation}
\begin{aligned}
    \mathbf S_{[i+1]}
    &= \mathbf S_{[i]}+\mathbf K_{[i+1]}^\top\mathbf V_{[i+1]},\\
    \mathbf O_{[i+1]}
    &= \mathbf Q_{[i+1]}\mathbf S_{[i]}
    +\left(\mathbf Q_{[i+1]}\mathbf K_{[i+1]}^\top\odot\mathbf M\right)\mathbf V_{[i+1]}.
\end{aligned}
\label{eq:background-chunkwise-linear-attention}
\end{equation}
The first output term retrieves from the state carried across the preceding chunks, while the second computes all causal interactions within the current chunk. Scalar decay preserves this structure through cumulative rescaling. Vector-valued forget gates and Delta Rule updates require more involved block computation \citep{yang2024gla,yang2024gateddelta}. \Cref{tab:memory-recurrences} summarizes the corresponding linear attention interfaces and their relative training costs.

\subsection{Slot-Based Memory}
\label{sec:background-slot-attention}

Slot-based memory retains softmax retrieval while compressing the prefix into $m$ key slots and $m$ value slots. Let $\mathbf K_t\in\mathbb R^{m\times d_k}$ and $\mathbf V_t\in\mathbb R^{m\times d_v}$. The readout is
\begin{equation}
    \boldsymbol o_t
    = \mathbf V_t^\top\operatorname{softmax}(\mathbf K_t\boldsymbol q_t).
    \label{eq:slot-readout}
\end{equation}
The methods below differ in how they allocate writes and erase existing content across the slots.

Sliding-window attention (SWA) stores the most recent $m$ key--value pairs exactly \citep{beltagy2020longformer}. Let $\boldsymbol e_i$ be the standard basis vector in $\mathbb R^m$ indexed by $i=0,\ldots,m-1$, and let $\mathbf Z=\sum_{i=0}^{m-2}\boldsymbol e_{i+1}\boldsymbol e_i^\top$ shift each row by one position toward the window boundary. SWA updates its states by
\begin{equation}
\begin{aligned}
    \mathbf K_t &= \mathbf Z\mathbf K_{t-1}+\boldsymbol e_0\boldsymbol k_t^\top,\\
    \mathbf V_t &= \mathbf Z\mathbf V_{t-1}+\boldsymbol e_0\boldsymbol v_t^\top.
\end{aligned}
\label{eq:background-swa-state}
\end{equation}
A key--value pair is discarded after it moves beyond the final slot. ABC replaces this fixed slot assignment with a learned write-allocation vector $\boldsymbol\phi_t\in\mathbb R^m$, allowing each key--value pair to be distributed across the memory slots \citep{peng2022abc}:
\begin{equation}
\begin{aligned}
    \mathbf K_t &= \mathbf K_{t-1}+\boldsymbol\phi_t\boldsymbol k_t^\top,\\
    \mathbf V_t &= \mathbf V_{t-1}+\boldsymbol\phi_t\boldsymbol v_t^\top.
\end{aligned}
\label{eq:background-abc-state}
\end{equation}
GSA adds a slot-wise forget gate $\boldsymbol\alpha_t\in(0,1)^m$, whose complement determines the write strength for each slot \citep{zhang2024gsa}:
\begin{equation}
\begin{aligned}
    \mathbf K_t
    &= \operatorname{Diag}(\boldsymbol\alpha_t)\mathbf K_{t-1}
     +(\boldsymbol 1-\boldsymbol\alpha_t)\boldsymbol k_t^\top,\\
    \mathbf V_t
    &= \operatorname{Diag}(\boldsymbol\alpha_t)\mathbf V_{t-1}
     +(\boldsymbol 1-\boldsymbol\alpha_t)\boldsymbol v_t^\top.
\end{aligned}
\label{eq:background-gsa-state}
\end{equation}

\paragraph{Two-pass computation.}
In ABC and GSA, the key and value states use the same slot-control signals. This shared structure allows the readout to be computed with two linear attention passes. The first pass matches the query against the key state to produce slot logits. A token-wise softmax converts these logits into slot weights, and the second pass retrieves from the value state using those weights:
\begin{equation}
\begin{array}{@{}c@{\quad}c@{}}
    \text{ABC} & \text{GSA} \\[0.4em]
    \begin{aligned}
        \{\boldsymbol o'_t\}_{t=1}^{T}
        &= \operatorname{LA}
           \bigl(\{\boldsymbol q_t,\boldsymbol k_t,\boldsymbol\phi_t\}_{t=1}^{T}\bigr),\\
        \boldsymbol o''_t
        &= \operatorname{softmax}(\boldsymbol o'_t),\\
        \{\boldsymbol o_t\}_{t=1}^{T}
        &= \operatorname{LA}
           \bigl(\{\boldsymbol o''_t,\boldsymbol\phi_t,\boldsymbol v_t\}_{t=1}^{T}\bigr)
    \end{aligned}
    &
    \begin{aligned}
        \{\boldsymbol o'_t\}_{t=1}^{T}
        &= \operatorname{ColumnGatedLA}
           \bigl(\{\boldsymbol q_t,\boldsymbol k_t,
           \boldsymbol 1-\boldsymbol\alpha_t,\boldsymbol\alpha_t\}_{t=1}^{T}\bigr),\\
        \boldsymbol o''_t
        &= \operatorname{softmax}(\boldsymbol o'_t),\\
        \{\boldsymbol o_t\}_{t=1}^{T}
        &= \operatorname{RowGatedLA}
           \bigl(\{\boldsymbol o''_t,
           \boldsymbol 1-\boldsymbol\alpha_t,\boldsymbol v_t,\boldsymbol\alpha_t\}_{t=1}^{T}\bigr)
    \end{aligned}
\end{array}
\label{eq:slot-two-pass}
\end{equation}
Because the softmax and the transformations between the two passes are token-wise, both recurrent calls can use hardware-efficient chunk-wise algorithms.

\section{Cyclic Flow Attention}
\label{sec:cyfa}

\subsection{Relative-Time-Partitioned Memory}
\label{sec:temporal_memory}

A fixed-size recurrent state must continually reuse its finite capacity as the
sequence grows. We propose to organize this reuse along a relative-time axis.
CyFA maintains key and value states whose rows correspond to relative-time
slots. Before the current key--value pair is written to the age-zero slot, we
transport the existing states toward larger model ages. Repeating this update
produces relative-time-partitioned memory within a fixed-size recurrent state.

Sliding-window attention (SWA) provides a discrete point of comparison for this
row-wise temporal organization. At each token, it shifts the existing rows by
one position before writing the current key--value pair to row zero. This
preserves the most recent pairs exactly, but it allocates one row to every token
and discards a pair when it reaches the window boundary. We retain the temporal
organization created by shifting before writing, while changing how the rows
are reused and how far the state moves at each token.

\paragraph{Cyclic transport.}
To reuse the rows without a fixed window boundary, we close SWA's one-way
shift into a cyclic permutation. For $m$ active relative-time slots, let
$\boldsymbol e_r\in\mathbb R^m$ denote the standard basis vector indexed by
$r=0,\ldots,m-1$. We write the two shift matrices side by side:
\begin{equation}
\begin{aligned}
    \mathbf Z
    &=\sum_{r=0}^{m-2}\boldsymbol e_{r+1}\boldsymbol e_r^\top
    =
    \begin{bmatrix}
        0      & 0      & \cdots & 0      & 0\\
        1      & 0      & \cdots & 0      & 0\\
        0      & 1      & \ddots & \vdots & \vdots\\
        \vdots & \ddots & \ddots & 0      & 0\\
        0      & \cdots & 0      & 1      & 0
    \end{bmatrix},
    \qquad
    \mathbf P
    &=\mathbf Z+\boldsymbol e_0\boldsymbol e_{m-1}^\top
    =
    \begin{bmatrix}
        0      & 0      & \cdots & 0      & 1\\
        1      & 0      & \cdots & 0      & 0\\
        0      & 1      & \ddots & \vdots & \vdots\\
        \vdots & \ddots & \ddots & 0      & 0\\
        0      & \cdots & 0      & 1      & 0
    \end{bmatrix}.
\end{aligned}
\label{eq:cyfa-cyclic-permutation}
\end{equation}
The matrices differ only in their top-right entry. The zero entry in
$\mathbf Z$ discards the final row and leaves the age-zero row empty for the
current write. Replacing it by one gives $\mathbf P$, which returns the final
row to the age-zero slot. This closes the one-way shift into cyclic transport
and reuses the same finite rows without a fixed window boundary. A unit shift,
however, still allocates one full row to every token. To decouple the
relative-time resolution from token positions, we extend $\mathbf P$ to
fractional cyclic shifts.

Because $\mathbf P$ is circulant, the discrete Fourier basis diagonalizes it
\citep{gray2006toeplitz}. We use an odd number of active slots $m$\footnote{For
even $m$, the real Fourier basis contains an additional unpaired Nyquist
component. Odd $m$ leaves one DC component and $(m-1)/2$ cosine--sine pairs.}
and let $\boldsymbol\Phi\in\mathbb R^{m\times m}$ denote the equivalent
orthogonal real basis. Its columns contain the normalized DC component followed
by cosine--sine pairs at frequencies $j=1,\ldots,(m-1)/2$. Let
\begin{equation}
\operatorname{Rot}(\theta)=
\begin{bmatrix}
    \cos\theta & -\sin\theta\\
    \sin\theta & \cos\theta
\end{bmatrix}.
\label{eq:cyfa-rotation-block}
\end{equation}
We then define
\begin{align}
    \mathcal U(\tau)
    &=
    1 \oplus
    \bigoplus_{j=1}^{(m-1)/2}
    \operatorname{Rot}\!\left(\frac{2\pi j\tau}{m}\right),
    \qquad
    \mathbf P(\tau)
    =
    \boldsymbol\Phi\mathcal U(\tau)\boldsymbol\Phi^\top,
    \qquad \tau\in\mathbb R .
    \label{eq:cyfa-fractional-shift}
\end{align}
This construction forms an orthogonal periodic family:
\begin{equation}
    \mathbf P(0)=\mathbf P(m)=\mathbf I,
    \qquad
    \mathbf P(1)=\mathbf P,
    \qquad
    \mathbf P(\tau_1)\mathbf P(\tau_2)=\mathbf P(\tau_1+\tau_2).
\label{eq:cyfa-shift-properties}
\end{equation}
The group law makes transport amounts additive across successive updates.
Orthogonality preserves the state norm and makes every shift reversible.
Consequently, $\mathbf P(\tau)$ continuously transports the state between
integer slots without introducing the boundary loss of SWA.

Not every token contributes the same amount of information that needs to remain
distinguishable in memory. A fixed fractional shift nevertheless advances the
states by the same amount at every step, placing successive writes at uniform
intervals along the relative-time axis. With a finite number of relative-time
slots, this gives the same temporal resolution to tokens that add little
information and to tokens whose content must remain clearly separated. We
therefore introduce a learned clock that allows different writes to receive
different relative-time intervals. At step $t$, we predict a clock increment
$\delta_t\in(0,1)$ and use it as the amount of cyclic transport applied before
the current write. A larger increment creates more separation between the
current write and preceding writes, whereas a smaller increment keeps
neighboring writes closer together. We accumulate these increments as
\begin{equation}
    \lambda_t=\lambda_{t-1}+\delta_t.
\label{eq:cyfa-cumulative-clock}
\end{equation}
For a write inserted at step $i$, the accumulated increment
$\lambda_t-\lambda_i$ defines its model age at step $t$.

Cyclic transport wraps contributions from the final relative-time slot back to
the age-zero slot, so their persistence is no longer determined by a fixed
window boundary. We therefore apply a scalar forget gate
$\alpha_t\in(0,1)$ \citep{dao2024mamba2,yang2024gateddelta} to gradually erase
existing content and use $\beta_t\in(0,1)$ as the write strength of the incoming
pair. Let $\mathbf K_t\in\mathbb R^{m\times d_k}$ and
$\mathbf V_t\in\mathbb R^{m\times d_v}$ denote the key and value states. We
update them as
\begin{equation}
\begin{aligned}
    \mathbf K_t
    &= \alpha_t\mathbf P(\delta_t)\mathbf K_{t-1}
     + \beta_t\boldsymbol e_0\boldsymbol k_t^\top,\\
    \mathbf V_t
    &= \alpha_t\mathbf P(\delta_t)\mathbf V_{t-1}
     + \beta_t\boldsymbol e_0\boldsymbol v_t^\top.
\end{aligned}
\label{eq:cyfa-physical-recurrence}
\end{equation}
The same cyclic transport and scalar forget gate act on both states, preserving
the alignment between each key contribution and its corresponding value
contribution. The current key--value pair is then written to the age-zero slot.

Unrolling the key-state recurrence shows how each earlier write is represented
at step $t$:
\begin{equation}
\mathbf K_t=\sum_{i\le t}\beta_i\!\left(\prod_{j=i+1}^{t}\alpha_j\right)
\mathbf P(\lambda_t-\lambda_i)\boldsymbol e_0\boldsymbol k_i^\top.
\label{eq:cyfa-state-expansion}
\end{equation}
For the write at step $i$, the product of forget gates determines its remaining
strength. Its relative-time profile at step $t$ is
\begin{equation}
    \mathbf P(\lambda_t-\lambda_i)\boldsymbol e_0.
\label{eq:cyfa-relative-time-profile}
\end{equation}
This profile determines how the contribution is distributed across the
relative-time slots according to its model age. Integer model ages recover
exact cyclic shifts, whereas fractional model ages produce their continuous
periodic interpolation. Because each later update applies the same orthogonal
transport to all existing contributions, their relative-time separation is
preserved. The value state has the same expansion with
$\boldsymbol v_i^\top$ in place of $\boldsymbol k_i^\top$.
Appendix~\ref{app:temporal-addressing} provides the detailed transport and
interpolation properties.

The key and value states therefore contain aligned contributions organized
across the same relative-time slots. To learn how these slots should be combined
during retrieval, we introduce a readout matrix
$\mathbf R\in\mathbb R^{m\times m}$ and apply it to both states. We initialize
$\mathbf R$ to the identity and compute
\begin{equation}
    \boldsymbol o_t
    =
    (\mathbf R\mathbf V_t)^\top
    \operatorname{softmax}\!\left(
        \mathbf R\mathbf K_t\boldsymbol q_t
    \right).
    \label{eq:cyfa-physical-readout}
\end{equation}
The readout matrix affects only retrieval. The recurrent states continue to
follow \cref{eq:cyfa-physical-recurrence}.\footnote{For simplicity, we omit the
conventional dot-product scaling factor from the notation throughout.}

\subsection{Scalar-Decay Recurrence in Absolute-Clock Coordinates}
\label{sec:scalar_decay_coordinates}

Directly evaluating the relative-time recurrence in
\cref{eq:cyfa-physical-recurrence} applies the dense $m\times m$ matrix
$\mathbf P(\delta_t)$ to both states at every token. This dense row mixing
precludes the compact matrix-multiply form required for efficient chunk-wise
computation \citep{yang2024gla}. We therefore derive an equivalent coordinate
representation that preserves the same memory update without repeatedly
shifting the stored matrices.

Because the fractional cyclic shift in \cref{eq:cyfa-fractional-shift} is
constructed in a real Fourier basis, we begin by expressing both states in
that basis:
\begin{equation}
    \widehat{\mathbf K}_t=\boldsymbol\Phi^\top\mathbf K_t,
    \qquad
    \widehat{\mathbf V}_t=\boldsymbol\Phi^\top\mathbf V_t,
    \qquad
    \boldsymbol b=\boldsymbol\Phi^\top\boldsymbol e_0.
\label{eq:cyfa-fourier-coordinates}
\end{equation}
In these coordinates, the relative-time recurrence in
\cref{eq:cyfa-physical-recurrence} becomes
\begin{equation}
\begin{aligned}
    \widehat{\mathbf K}_t
    &= \alpha_t\mathcal U(\delta_t)\widehat{\mathbf K}_{t-1}
     + \beta_t\boldsymbol b\boldsymbol k_t^\top,\\
    \widehat{\mathbf V}_t
    &= \alpha_t\mathcal U(\delta_t)\widehat{\mathbf V}_{t-1}
     + \beta_t\boldsymbol b\boldsymbol v_t^\top.
\end{aligned}
\label{eq:cyfa-relative-fourier}
\end{equation}
Dense mixing among the relative-time slots is now replaced by independent
two-dimensional rotations of the Fourier frequency pairs. However,
$\mathcal U(\delta_t)$ still depends on the current token and remains inside
the recurrent transition.

Block diagonalization therefore resolves only the dense row mixing. To remove
the remaining rotations from the recurrent transition, we let the coordinates
track the cumulative clock and define the absolute-clock states by
\begin{equation}
    \overline{\mathbf K}_t
    =\mathcal U(-\lambda_t)\widehat{\mathbf K}_t,
    \qquad
    \overline{\mathbf V}_t
    =\mathcal U(-\lambda_t)\widehat{\mathbf V}_t.
\label{eq:cyfa-absolute-coordinates}
\end{equation}
Since $\lambda_t=\lambda_{t-1}+\delta_t$, the group law in
\cref{eq:cyfa-shift-properties} yields
\begin{equation}
    \mathcal U(-\lambda_t)\mathcal U(\delta_t)
    =\mathcal U(-\lambda_{t-1}).
\label{eq:cyfa-clock-cancellation}
\end{equation}
Under the same change of coordinates, the fixed age-zero insertion is
represented by the temporal write vector
$\mathcal U(-\lambda_t)\boldsymbol b$. Applying the identity in
\cref{eq:cyfa-clock-cancellation} to the Fourier-coordinate recurrence gives
\begin{equation}
\begin{aligned}
    \overline{\mathbf K}_t
    &= \alpha_t\overline{\mathbf K}_{t-1}
     + \beta_t\mathcal U(-\lambda_t)\boldsymbol b\boldsymbol k_t^\top,\\
    \overline{\mathbf V}_t
    &= \alpha_t\overline{\mathbf V}_{t-1}
     + \beta_t\mathcal U(-\lambda_t)\boldsymbol b\boldsymbol v_t^\top.
\end{aligned}
\label{eq:cyfa-absolute-recurrence}
\end{equation}
Equation~\eqref{eq:cyfa-absolute-recurrence} is the key computational
consequence of absolute-clock coordinates. The learned clock controls how each
new write enters through the temporal write vector, while the stored key and
value states are carried forward only through the scalar decay $\alpha_t$.

The change of coordinates preserves the original memory update exactly.
Reconstructing the previous key state and the temporal write vector at clock
value $\lambda_t$ gives
\begin{equation}
\begin{aligned}
    \boldsymbol\Phi\mathcal U(\lambda_t)\overline{\mathbf K}_{t-1}
    &=\mathbf P(\delta_t)\mathbf K_{t-1},
    &
    \boldsymbol\Phi\mathcal U(\lambda_t)
    \mathcal U(-\lambda_t)\boldsymbol b
    &=\boldsymbol e_0.
\end{aligned}
\label{eq:cyfa-transport-correspondence}
\end{equation}
The first identity recovers the cyclic transport of the previous key state,
and the second maps the temporal write vector to the age-zero slot. Together,
they show that \cref{eq:cyfa-absolute-recurrence} represents the same
relative-time memory update as \cref{eq:cyfa-physical-recurrence}. The value
state follows the same correspondence.

For retrieval, the current clock maps the absolute-clock states back to
relative-time coordinates:
\begin{equation}
    \mathbf K_t
    =\boldsymbol\Phi\mathcal U(\lambda_t)\overline{\mathbf K}_t,
    \qquad
    \mathbf V_t
    =\boldsymbol\Phi\mathcal U(\lambda_t)\overline{\mathbf V}_t.
\label{eq:cyfa-relative-reconstruction}
\end{equation}
Substituting \cref{eq:cyfa-relative-reconstruction} into the softmax readout in
\cref{eq:cyfa-physical-readout} and reassociating the matrix products gives
\begin{equation}
    \boldsymbol o_t
    =
    \overline{\mathbf V}_t^\top
    \mathcal U(-\lambda_t)\boldsymbol\Phi^\top\mathbf R^\top
    \operatorname{softmax}\!\left(
        \mathbf R\boldsymbol\Phi\mathcal U(\lambda_t)
        \overline{\mathbf K}_t\boldsymbol q_t
    \right).
    \label{eq:cyfa-absolute-readout}
\end{equation}
After reassociation, the clock-dependent transformations in
\cref{eq:cyfa-absolute-readout} act on token-wise $m$-dimensional vectors rather
than on the stored matrices. The recurrent key and value states continue to
follow the scalar-decay updates in \cref{eq:cyfa-absolute-recurrence}. This
separation leads directly to the two-pass chunk-wise computation in the next
section.

\subsection{Two-Pass Chunk-Wise Computation}
\label{sec:two_pass_cyfa}

The recurrence in \cref{eq:cyfa-absolute-recurrence} has the scalar-decay
rank-one form supported by the $\operatorname{ScalarGatedLA}$ interface in
\cref{tab:memory-recurrences}. We therefore implement the key and value states
as two recurrent passes connected by a token-wise softmax readout:
\begin{equation}
\begin{aligned}
    \{\boldsymbol o'_t\}_{t=1}^{T}
    &=
    \operatorname{ScalarGatedLA}
    \left(
        \left\{
        \boldsymbol q_t,\,
        \boldsymbol k_t,\,
        \beta_t\mathcal U(-\lambda_t)\boldsymbol b,\,
        \alpha_t
        \right\}_{t=1}^{T}
    \right),\\
    \boldsymbol o''_t
    &=
    \mathcal U(-\lambda_t)\boldsymbol\Phi^\top\mathbf R^\top
    \operatorname{softmax}\!\left(
        \mathbf R\boldsymbol\Phi\mathcal U(\lambda_t)\boldsymbol o'_t
    \right),\\
    \{\boldsymbol o_t\}_{t=1}^{T}
    &=
    \operatorname{ScalarGatedLA}
    \left(
        \left\{
        \boldsymbol o''_t,\,
        \beta_t\mathcal U(-\lambda_t)\boldsymbol b,\,
        \boldsymbol v_t,\,
        \alpha_t
        \right\}_{t=1}^{T}
    \right).
\end{aligned}
\label{eq:cyfa-two-pass}
\end{equation}
The key pass produces
$\boldsymbol o'_t=\overline{\mathbf K}_t\boldsymbol q_t$, an $m$-dimensional
contraction of the query with the stored key state. The middle line maps this
vector back to relative-time coordinates, applies the readout matrix to obtain
the slot logits, normalizes them into slot weights, and maps those weights back
to absolute-clock coordinates. The resulting vector $\boldsymbol o''_t$ then
queries the stored value state in the value pass.

All computation between the two recurrent passes is token-wise and acts only on
$m$-dimensional vectors. The cumulative clock sequence
$\{\lambda_t\}_{t=1}^{T}$ is obtained from the clock increments
$\{\delta_t\}_{t=1}^{T}$ by a parallel prefix sum. Both recurrent passes can
therefore use existing hardware-efficient chunk-wise algorithms for
$\operatorname{ScalarGatedLA}$ \citep{dao2024mamba2,yang2024fla}, without
materializing either relative-time state at every token. A detailed formulation
is given in \cref{app:chunkwise_parallel}.

\paragraph{Relation to Blurry Window Attention.}
With a fixed-rate clock $\delta_t\equiv1/\rho$, CyFA's relative-time profile
specializes to the Fourier--Dirichlet profile used by Blurry Window Attention
\citep{laborieux2026bla}, as derived in Appendix~\ref{app:bla}.

\subsection{Network Design}
\label{sec:network_design}

\paragraph{Per-head parameterization.}
Each CyFA head receives the layer input $\boldsymbol x_t$ and uses learned
projections to produce $\boldsymbol q_t^h$, $\boldsymbol k_t^h$, and
$\boldsymbol v_t^h$, together with the clock increment $\delta_t^h$, scalar
forget gate $\alpha_t^h$, and write strength $\beta_t^h$. Each head also has a
learned readout matrix $\mathbf R^h$, initialized to the identity.
In our implementation, each head uses $m=127$ active slots. We allocate
$m^\ast=128$ storage rows for hardware-efficient computation and mask the
padding row during writing and readout. Suppressing the head index, we
parameterize the scalar forget gate following prior Linear RNNs
\citep{dao2024mamba2,yang2024gateddelta} as
$\alpha_t=\exp\!\left(-A\,\operatorname{softplus}\!\left(
\mathbf W_{\alpha}\boldsymbol x_t+\boldsymbol b_{\alpha}\right)\right)$.

\begin{figure}[t]
    \centering
    \begin{minipage}[t]{0.59\linewidth}
        \vspace{0pt}
        \centering
        \includegraphics[width=\linewidth]{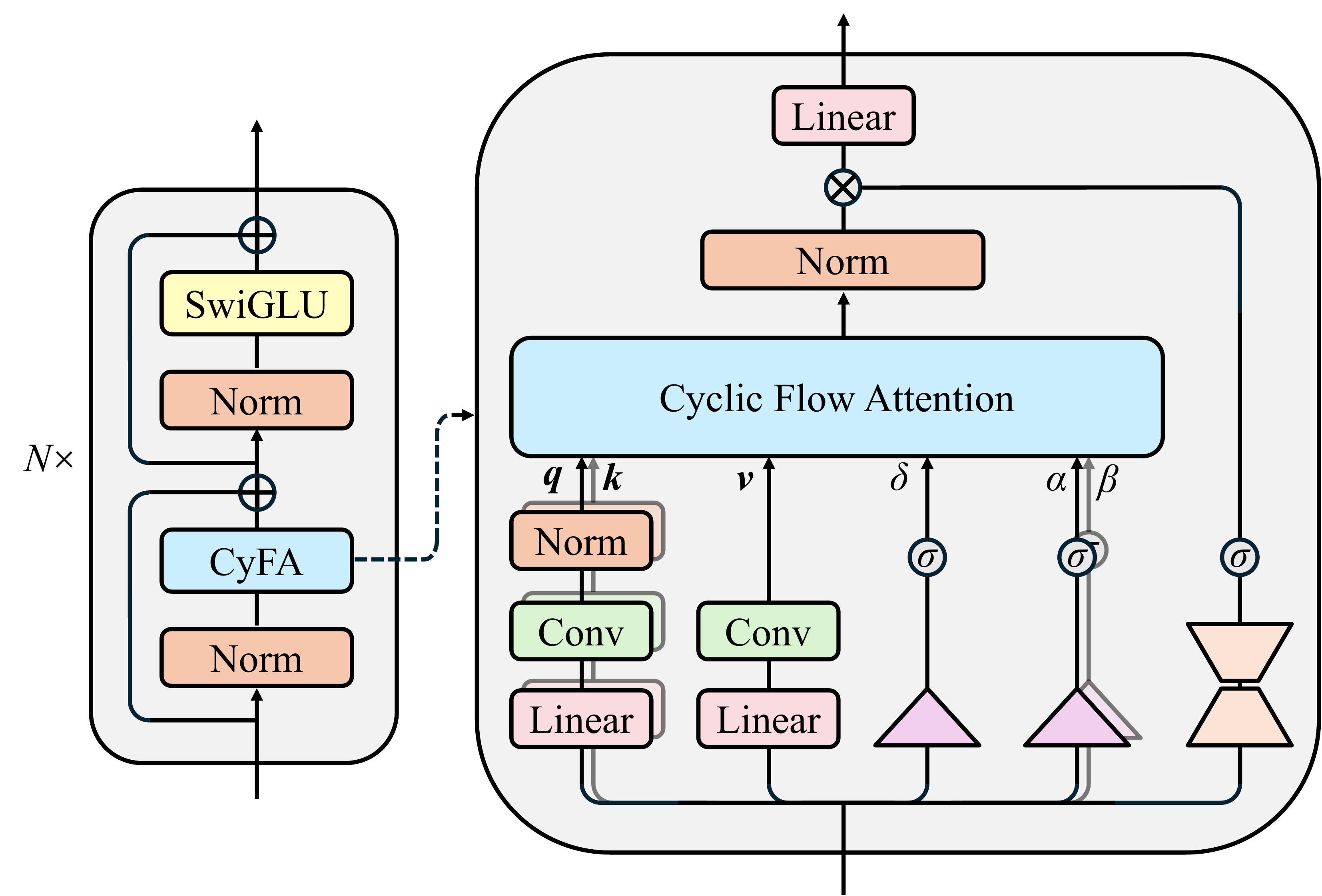}
        \captionof{figure}{Overall CyFA architecture. The left panel shows the pre-norm backbone, and the right panel expands one CyFA mixer.}
        \label{fig:cyfa-architecture}
    \end{minipage}\hfill
    \begin{minipage}[t]{0.39\linewidth}
        \vspace{0pt}
        \centering
        \includegraphics[width=\linewidth]{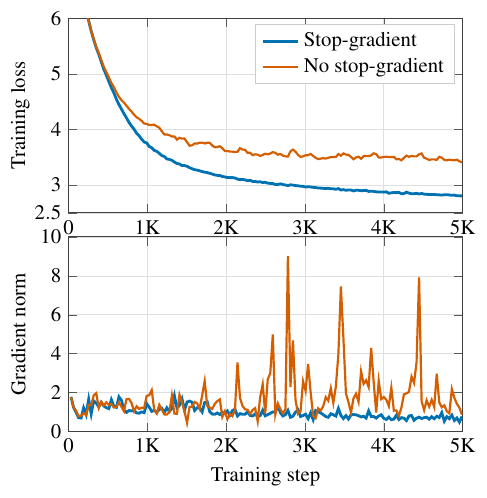}
        \captionof{figure}{Clock gradient stabilization. Training loss and gradient norm during the first 5,000 steps, with and without stop-gradient in the clock-increment projection.}
        \label{fig:clock-gradient-stabilization}
    \end{minipage}
\end{figure}

\paragraph{Mixer and backbone.}
Within each CyFA mixer, ShortConv is applied to $\boldsymbol q$, $\boldsymbol k$,
and $\boldsymbol v$
\citep{gu2023mamba,dao2024mamba2,yang2024gateddelta,kimiteam2025kimi}.
RMSNorm then normalizes the queries and keys
\citep{yang2025qwen35}. The multi-head outputs pass through a head-wise RMSNorm
\citep{qin2022devil,lan2025liger,yang2024deltanetparallel,yang2024gateddelta,kimiteam2025kimi}
and a low-rank sigmoid output gate
\citep{qiu2025gatedattention,kimiteam2025kimi}. At the backbone level, CyFA
replaces self-attention in a pre-norm LLaMA-style architecture
\citep{touvron2023llama} and alternates with SwiGLU feed-forward layers
\citep{shazeer2020glu}. Figure~\ref{fig:cyfa-architecture} summarizes the
resulting model.

\paragraph{Clock gradient stabilization.}
Each clock increment affects subsequent positions through the cumulative clock.
Backpropagation therefore connects the clock-increment branch to later temporal
write vectors and readouts through a long gradient path. In practice, we
observed persistent loss oscillations and repeated gradient spikes when
gradients were allowed to propagate through this path.

We stabilize the learned clock by stopping the gradient from the clock-increment
projection into its layer input:
\begin{equation}
    \delta_t^h
    =
    \sigma\!\left(
        (\mathbf W_{\delta}\operatorname{sg}(\boldsymbol x_t)
        +\boldsymbol b_{\delta})_h
    \right),
    \label{eq:clock_gradient_stabilization}
\end{equation}
The stop-gradient operator $\operatorname{sg}$ leaves its forward value
unchanged and has zero derivative. Gradients through the temporal write vectors
and token-wise readout continue to train $\mathbf W_{\delta}$ and
$\boldsymbol b_{\delta}$, while the clock-increment branch no longer propagates
cumulative-clock gradients into $\boldsymbol x_t$.
Figure~\ref{fig:clock-gradient-stabilization} shows that this intervention
produces a smooth decrease in training loss and stable gradient norms. Removing
stop-gradient results in persistent loss oscillations and repeated gradient
spikes after the initial descent. \Cref{app:clock_gradient_analysis} analyzes
the underlying gradient path.

\begin{table}[!t]
    \centering
    \caption{Language modeling, zero-shot commonsense reasoning, and recall-intensive evaluation. Commonsense tasks are evaluated with \texttt{lm-evaluation-harness} \citep{gao2023evalharness}. Recall-intensive tasks are evaluated with \texttt{prefix-linear-attention} \citep{arora2024justreadtwice}, with inputs truncated to 2K tokens. Public checkpoints in the 1.4B reference group are marked with $^{\dagger}$.}
    \label{tab:language_recall_results}
    \scriptsize
    \setlength{\tabcolsep}{4pt}
    \renewcommand{\arraystretch}{1.12}
    \begin{cypawidetable}
    \begin{tabular}{l|cc|ccccccc|ccccccc}
        \toprule
        \multirow[c]{2}{*}[-0.7ex]{\textbf{Model}} &
        \multicolumn{2}{c|}{Perplexity} &
        \multicolumn{7}{c|}{Commonsense Reasoning Tasks} &
        \multicolumn{7}{c}{Recall-Intensive Tasks} \\
        \cmidrule{2-17}
        &
        \makecell{\textbf{Wiki.}\\ppl$\downarrow$} &
        \makecell{\textbf{Lamb.}\\ppl$\downarrow$} &
        \makecell{\textbf{ARC-e}\\acc$\uparrow$} &
        \makecell{\textbf{ARC-c}\\acc$_n\uparrow$} &
        \makecell{\textbf{Hella.}\\acc$_n\uparrow$} &
        \makecell{\textbf{Lamb.}\\acc$\uparrow$} &
        \makecell{\textbf{PIQA}\\acc$\uparrow$} &
        \makecell{\textbf{Wino.}\\acc$\uparrow$} &
        \makecell{\textbf{Avg.}\\acc$\uparrow$} &
        \makecell{\textbf{FDA}\\acc$\uparrow$} &
        \makecell{\textbf{SWDE}\\acc$\uparrow$} &
        \makecell{\textbf{SQD}\\acc$\uparrow$} &
        \makecell{\textbf{NQ}\\acc$\uparrow$} &
        \makecell{\textbf{TQA}\\acc$\uparrow$} &
        \makecell{\textbf{DROP}\\acc$\uparrow$} &
        \makecell{\textbf{Avg.}\\acc$\uparrow$} \\
        \midrule
        \multicolumn{17}{l}{\textit{400M parameters with 15B training tokens, $L=24$, and $d=1024$}} \\
        \midrule
        \rowcolor{black!6}
        Transformer & 27.30 & 74.28 & 44.74 & 24.23 & 35.25 & 26.02 & 64.69 & 51.46 & 41.07 & 53.04 & 38.89 & 30.70 & 16.41 & 45.68 & 19.17 & 33.98 \\
        SWA         & 33.29 & 89.22 & 46.13 & 22.61 & 34.52 & 25.05 & 63.71 & 49.33 & 40.23 & 15.62 & 8.72 & 21.46 & 5.29 & 31.52 & 16.20 & 16.47 \\
        GSA         & 28.56 & 87.55 & 46.25 & \textbf{25.00} & 34.40 & 23.85 & 64.80 & 51.78 & 41.01 & 5.54 & 14.15 & 23.65 & 10.90 & 41.05 & 17.54 & 18.81 \\
        Raven       & 32.14 & 91.49 & 43.81 & 23.21 & 32.58 & 24.22 & 64.42 & 50.91 & 39.86 & 19.80 & 25.77 & 25.73 & 9.76 & 40.17 & 15.86 & 22.85 \\
        BLA ($\rho=2$) & 32.88 & 79.86 & 44.49 & 22.78 & 33.90 & 25.21 & 64.31 & 50.75 & 40.24 & 12.62 & 11.72 & 28.08 & 5.67 & 38.86 & 16.63 & 18.93 \\
        BLA ($\rho=5$) & 33.87 & 173.33 & 43.52 & 23.98 & 32.18 & 19.02 & 63.71 & 51.30 & 38.95 & 5.99 & 16.40 & 23.55 & 6.72 & 39.63 & 14.37 & 17.78 \\
        Mamba-2     & 28.25 & 89.36 & 45.75 & 22.61 & 34.95 & 22.34 & 63.98 & 52.33 & 40.33 & 9.08 & 21.18 & 25.13 & 11.97 & 40.23 & 16.10 & 20.62 \\
        Mamba-3     & 26.75 & \underline{49.99} & \underline{46.46} & 22.70 & 36.16 & \underline{27.96} & 64.31 & \underline{53.12} & 41.79 & 24.89 & \underline{27.84} & \underline{29.53} & 14.92 & 43.84 & 17.73 & \underline{26.46} \\
        GDN         & 27.10 & 62.92 & 43.94 & 23.21 & 35.25 & 25.91 & 64.42 & 51.78 & 40.75 & 17.35 & 23.81 & 27.31 & \underline{15.11} & 42.36 & \underline{18.78} & 24.12 \\
        KDA         & \textbf{25.76} & 51.17 & \textbf{47.10} & 23.21 & \underline{36.50} & 27.36 & \textbf{65.51} & 51.70 & \underline{41.90} & \underline{26.07} & 25.02 & 29.49 & 14.82 & \underline{45.38} & 17.06 & 26.31 \\
        CyFA        & \underline{25.85} & \textbf{48.17} & 45.92 & \underline{24.06} & \textbf{36.84} & \textbf{29.26} & \underline{65.29} & \textbf{53.43} & \textbf{42.47} & \textbf{42.60} & \textbf{35.61} & \textbf{34.63} & \textbf{17.33} & \textbf{48.70} & \textbf{20.65} & \textbf{33.25} \\
        \midrule
        \multicolumn{17}{l}{\textit{800M parameters with 30B training tokens, $L=24$, and $d=1536$}} \\
        \midrule
        \rowcolor{black!6}
        Transformer  & 20.41 & 23.32 & 50.29 & 24.83 & 42.97 & 36.99 & 67.90 & 50.67 & 45.61 & 49.32 & 42.92 & 39.17 & 22.24 & 54.44 & 21.27 & 38.23 \\
        GDN          & 20.44 & 25.42 & 52.02 & \textbf{27.05} & 42.87 & 36.06 & 68.50 & \underline{53.75} & 46.71 & \underline{34.15} & 28.58 & 32.65 & 18.37 & 52.61 & \underline{19.26} & 30.94 \\
        KDA          & \underline{19.83} & \underline{22.20} & \underline{52.36} & 25.51 & \underline{44.03} & \underline{36.76} & \underline{68.66} & 53.51 & \underline{46.81} & 28.25 & \underline{30.18} & \underline{33.93} & \underline{21.95} & \underline{53.55} & 19.21 & \underline{31.18} \\
        CyFA         & \textbf{19.51} & \textbf{18.93} & \textbf{53.62} & \underline{26.11} & \textbf{44.65} & \textbf{39.74} & \textbf{70.13} & \textbf{54.70} & \textbf{48.16} & \textbf{43.69} & \textbf{40.58} & \textbf{39.03} & \textbf{24.17} & \textbf{55.21} & \textbf{20.12} & \textbf{37.13} \\
        \midrule
        \multicolumn{17}{l}{\textit{1.4B parameters with 100B training tokens, $L=24$, and $d=2048$}} \\
        \midrule
        \rowcolor{black!6}
        Transformer$^{\dagger}$ & 17.65 & 18.61 & 56.02 & 28.33 & 49.11 & 40.95 & 69.86 & 54.85 & 49.85 & 55.31 & 44.70 & 43.10 & 24.58 & 59.12 & 21.66 & 41.41 \\
        RetNet$^{\dagger}$     & 18.23 & 23.19 & 57.11 & 26.88 & 48.08 & 37.36 & 68.88 & 54.22 & 48.76 & 20.98 & 27.18 & 33.79 & 15.55 & 53.44 & 19.65 & 28.43 \\
        GLA$^{\dagger}$        & 17.68 & 19.72 & 55.18 & 27.30 & 48.80 & 40.33 & 70.02 & 52.96 & 49.10 & 27.25 & 31.12 & 34.73 & 22.27 & 55.45 & 19.26 & 31.68 \\
        GSA$^{\dagger}$        & 16.81 & 15.69 & \underline{58.75} & \underline{28.33} & 50.96 & 41.98 & \underline{71.93} & 52.72 & 50.78 & 23.89 & 29.99 & 35.98 & 23.15 & 57.46 & 20.89 & 31.89 \\
        KDA          & \underline{15.97} & \textbf{11.88} & \underline{58.75} & \textbf{28.41} & \underline{53.85} & \textbf{47.76} & 71.87 & \underline{56.67} & \underline{52.89} & \underline{42.60} & \underline{40.77} & \underline{37.79} & \underline{24.83} & \underline{58.89} & \underline{21.47} & \underline{37.73} \\
        CyFA         & \textbf{15.47} & \underline{12.77} & \textbf{60.98} & 28.24 & \textbf{54.32} & \underline{46.38} & \textbf{72.09} & \textbf{57.14} & \textbf{53.19} & \textbf{57.77} & \textbf{46.58} & \textbf{42.83} & \textbf{29.87} & \textbf{61.49} & \textbf{22.28} & \textbf{43.47} \\
        \bottomrule
    \end{tabular}
    \end{cypawidetable}
\end{table}

\section{Experiments}

\paragraph{Experimental setup.}
We compare CyFA with a LLaMA-style Transformer and eight efficient baselines:
SWA, GSA, Raven, BLA, Mamba-2, Mamba-3, GDN, and KDA
\citep{vaswani2017attention,touvron2023llama,beltagy2020longformer,zhang2024gsa,afzal2026raven,laborieux2026bla,dao2024mamba2,lahoti2026mamba3,yang2024gateddelta,kimiteam2025kimi}.

We train every architecture at approximately 400M parameters and extend
selected comparisons to 800M and 1.4B.

Within each scale, we adjust the number of heads and their state dimensions so that all recurrent models have the same total main recurrent state size. At 1.4B, we additionally report publicly released checkpoints from the FLA model collection on Hugging Face\footnote{\url{https://huggingface.co/fla-hub}} \citep{yang2024fla} as a separate reference group. These checkpoints retain their released configurations and are therefore outside the state-matched comparison. \Cref{app:architecture_state_matching} provides the exact architectures and state-size accounting.

All models trained in this study use a 24-layer pre-norm LLaMA-style backbone and are implemented with the \texttt{flash-linear-attention} library \citep{yang2024fla}. We pretrain on SlimPajama-627B \citep{soboleva2023slimpajama} with the Mistral tokenizer \citep{jiang2023mistral} and an initial context length of 2,048 tokens. The 400M, 800M, and 1.4B models receive 15B, 30B, and 100B training tokens, respectively, with global batches of 0.5M tokens at the first two scales and 1M tokens at 1.4B. We use AdamW with a peak learning rate of $3\times10^{-4}$, a weight decay of $0.01$ \citep{loshchilov2019adamw}, gradient clipping at $1.0$, and 1,024 warmup steps followed by cosine decay to $3\times10^{-5}$ \citep{loshchilov2017sgdr}. For the long-context evaluations in \cref{sec:long_context_ability}, we continue pretraining the 800M and 1.4B models for 1,024 steps at a context length of 8,192 while preserving the number of tokens per optimizer step. Training runs use eight NVIDIA RTX Pro 6000 GPUs. \Cref{app:experimental_details} provides the remaining experimental details.

\subsection{Language Modeling}
\label{sec:language_modeling}
\raggedbottom

\paragraph{Commonsense Reasoning.}

Following prior work \citep{zhang2024gsa}, we evaluate zero-shot performance on ARC-Easy (ARC-e) and ARC-Challenge (ARC-c) \citep{clark2018arc}, HellaSwag (Hella.) \citep{zellers2019hellaswag}, LAMBADA (Lamb.) \citep{paperno2016lambada}, PIQA \citep{bisk2020piqa}, and WinoGrande (Wino.) \citep{sakaguchi2019winogrande}. These benchmarks mostly use short inputs and therefore place limited demands on in-context retrieval. They instead provide a standard evaluation of general language understanding. As shown in \cref{tab:language_recall_results}, CyFA achieves the highest average accuracy at every model scale.

\paragraph{Recall-Intensive Tasks.}

Following the protocol of \citet{arora2024justreadtwice}, we evaluate real-world recall-intensive tasks including FDA \citep{arora2023evaporate}, SWDE \citep{arora2023evaporate,lockard2019opencere}, SQD \citep{rajpurkar2018squad}, NQ \citep{kwiatkowski2019naturalquestions}, TQA \citep{joshi2017triviaqa}, and DROP \citep{dua2019drop}. These benchmarks are particularly challenging for recurrent models because the information needed to answer each query must be recovered from a compressed recurrent state. Across all three model scales, CyFA is the strongest recurrent model on every task. The largest margins occur on FDA and SWDE, which require retrieving a target value from many coexisting field--value associations. At 400M, CyFA exceeds the next-best recurrent model by 16.53 points on FDA and 7.77 points on SWDE. At 1.4B, CyFA reaches an average recall score of 43.47, compared with 37.73 for KDA and 41.41 for the public Transformer reference.

\flushbottom

\subsection{Long-Context Ability}
\label{sec:long_context_ability}

All results in this section use models continued for 1,024 steps at an 8K context length. \Cref{app:additional_experimental_results} reports results before continuation.

\paragraph{Long-Context Understanding.}

We evaluate long-context understanding on LongBench \citep{bai2024longbench}. CyFA leads both code tasks at 800M and 1.4B, with its largest code gain on LCC at 1.4B (53.60 vs. 46.34 for KDA). It also achieves the highest average score at both scales, leading 10 of the 15 tasks at 800M and 11 of the 15 tasks at 1.4B.

\begin{table}[H]
    \centering
    \caption{LongBench results after 1,024-step continuation at an 8K training context, with inputs truncated to 8K tokens.}
    \label{tab:longbench_after_8k}
    \cypatablefont
    \setlength{\tabcolsep}{2.8pt}
    \renewcommand{\arraystretch}{1.05}
    \resizebox{\linewidth}{!}{%
    \begin{tabular}{l|l|cc|ccc|ccc|ccc|ccc|c}
        \toprule
        \multirow[c]{2}{*}[-0.4ex]{\textbf{Scale}} & \multirow[c]{2}{*}[-0.4ex]{\textbf{Model}} & \multicolumn{2}{c|}{Code} & \multicolumn{3}{c|}{Summarization} & \multicolumn{3}{c|}{SingleQA} & \multicolumn{3}{c|}{MultiQA} & \multicolumn{3}{c|}{Few-Shot} & \multirow{2}{*}{\textbf{Avg.}} \\
        \cmidrule{3-16}
        & & \textbf{LCC} & \textbf{RBP} & \textbf{GvR} & \textbf{QMS} & \textbf{MNs} & \textbf{NQA} & \textbf{QQA} & \textbf{MQA} & \textbf{HQA} & \textbf{2WM} & \textbf{MSQ} & \textbf{TRE} & \textbf{TQA} & \textbf{SAM} & \\
        \midrule
        \multirow{4}{*}{800M} & Transformer & 42.46 & 37.20 & 0.68 & 11.41 & 0.56 & 1.54 & 3.70 & 6.32 & 3.58 & 7.97 & 2.40 & 16.50 & \underline{26.21} & 5.50 & 11.86 \\
        & GDN & \underline{43.65} & 35.64 & \textbf{5.93} & 10.95 & 2.14 & \underline{3.32} & 1.67 & \underline{11.58} & \textbf{5.02} & \underline{8.79} & \underline{2.62} & 31.00 & 26.04 & \textbf{15.61} & 14.57 \\
        & KDA & 41.47 & \underline{39.06} & \underline{5.39} & \underline{11.54} & \underline{2.22} & 2.72 & \underline{5.29} & 10.88 & 3.90 & 8.12 & 2.42 & \textbf{47.50} & 25.55 & 10.47 & \underline{15.47} \\
        & CyFA & \textbf{47.84} & \textbf{43.59} & 3.54 & \textbf{15.43} & \textbf{5.57} & \textbf{3.50} & \textbf{6.17} & \textbf{14.03} & \underline{4.63} & \textbf{10.27} & \textbf{2.87} & \underline{32.50} & \textbf{37.78} & \underline{12.90} & \textbf{17.19} \\
        \midrule
        \multirow{2}{*}{1.4B} & KDA & \underline{46.34} & \underline{42.56} & \underline{4.26} & \underline{13.75} & \underline{7.68} & \underline{1.97} & \underline{1.55} & \underline{7.93} & \textbf{6.40} & \underline{8.66} & \underline{3.54} & \textbf{57.50} & \underline{49.09} & \textbf{29.11} & \underline{20.02} \\
        & CyFA & \textbf{53.60} & \textbf{46.40} & \textbf{5.98} & \textbf{14.50} & \textbf{11.58} & \textbf{2.82} & \textbf{5.71} & \textbf{14.85} & \underline{5.84} & \textbf{10.08} & \textbf{4.88} & \underline{43.50} & \textbf{56.37} & \underline{26.91} & \textbf{21.64} \\
        \bottomrule
    \end{tabular}}
\end{table}

\paragraph{Needle-In-A-Haystack (NIAH).}

We evaluate long-context retrieval with RULER \citep{hsieh2024ruler}. On S-NIAH-1/2/3, recurrent models remain competitive because each example requires retaining only one queried association. MK-NIAH-1 is more revealing: multiple key--value associations must coexist in the fixed-size recurrent state, making interference among them a central difficulty. GDN and KDA already suffer large accuracy drops at 1K--2K, whereas CyFA degrades more gradually and remains clearly ahead through 4K. This contrast supports CyFA's central premise that organizing memory along relative time improves retrieval when multiple associations must be preserved simultaneously.

\begin{figure}[H]
    \centering
    \includegraphics[width=\linewidth]{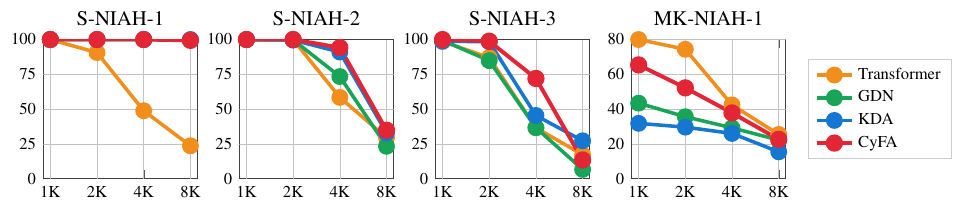}
    \caption{RULER accuracy on three single-needle tasks (S-NIAH-1/2/3) and one multi-key task (MK-NIAH-1) for 800M models after 1,024-step continuation at an 8K training context. Evaluation sequence lengths range from 1K to 8K tokens.}
    \label{fig:ruler_after_continuation}
\end{figure}

\subsection{Efficiency}

\begin{figure}[!ht]
    \centering
    \includegraphics[width=0.333333\linewidth]{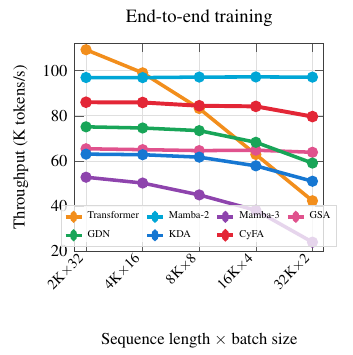}%
    \includegraphics[width=0.666667\linewidth]{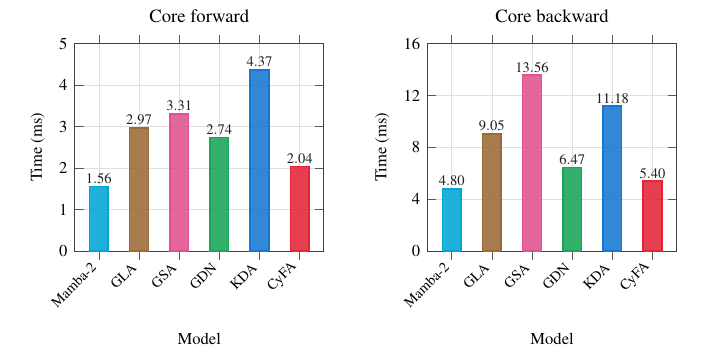}
    \par
    \begin{minipage}[t]{0.32\linewidth}
        \vspace{0pt}
        \centering
        \captionof{figure}{End-to-end training throughput of 400M models on a single NVIDIA RTX Pro 6000.}
        \label{fig:efficiency_400m}
    \end{minipage}\hfill
    \begin{minipage}[t]{0.66\linewidth}
        \vspace{0pt}
        \centering
        \captionof{figure}{Core operator execution times on a single NVIDIA RTX Pro 6000 using the 400M training shapes, a 2K sequence length, and batch size 32. GLA uses the same shapes as GDN and KDA.}
        \label{fig:kernel_latency_400m}
    \end{minipage}
\end{figure}

\Cref{fig:efficiency_400m} compares end-to-end training throughput while keeping the number of tokens per batch fixed. Because Raven and BLA reuse GSA's recurrent computation, we omit them from the figure. CyFA maintains approximately 80--86K tokens/s from 2K to 32K sequences, surpasses the Transformer from 8K onward, and is $1.37\times$ faster than KDA at 8K. Among the recurrent baselines, CyFA trails only Mamba-2, which uses a single-pass scalar recurrence rather than CyFA's two passes. The two-pass scalar-decay formulation therefore realizes relative-time-partitioned memory while retaining high training throughput.

We also benchmark the forward and backward execution times of the core operators (\cref{fig:kernel_latency_400m}). With the 2K sequence length and batch size 32 used for 400M training, CyFA's forward and backward passes take only $46.7\%$ and $48.3\%$ of KDA's execution time, respectively.

\subsection{Ablations}

\begingroup
\setlength{\intextsep}{0pt}
\begin{wraptable}{r}{0.44\linewidth}
        \vspace{-2\baselineskip}
        \centering
        \caption{CyFA ablations at 400M scale.}
        \label{tab:ablations}
        \cypatablefont
        \setlength{\tabcolsep}{4pt}
        \renewcommand{\arraystretch}{1.10}
        \resizebox{\linewidth}{!}{%
        \begin{tabular}{lcccc}
            \toprule
            \textbf{Variant} &
            \makecell{\textbf{Wiki.}\\ppl$\downarrow$} &
            \makecell{\textbf{Lamb.}\\ppl$\downarrow$} &
            \makecell{\textbf{Common}\\Avg.$\uparrow$} &
            \makecell{\textbf{Recall}\\Avg.$\uparrow$} \\
            \midrule
            CyFA w. $m = 127$          & 25.85 & 48.17 & 42.47 & 33.25 \\
            \midrule
            \multicolumn{5}{l}{\textit{Method}} \\
            \quad w. $\alpha_t = 1$         & 30.93 & 115.89 & 39.61 & 28.05 \\
            \quad w. $\delta_t^h\equiv\sigma(b_\delta^h)$ & 26.40 & 51.98 & 42.15 & 30.32 \\
            \quad w. $\mathbf R=\mathbf I$  & 26.26 & 58.75 & 41.89 & 31.79 \\
            \midrule
            \multicolumn{5}{l}{\textit{Architecture}} \\
            \quad w/o. ShortConv                & 26.34 & 52.05 & 42.04 & 31.70 \\
            \quad w/o. QK Norm                  & 26.29 & 51.37 & 42.61 & 30.12 \\
            \quad w. SiLU                       & 25.77 & 45.11 & 42.50 & 32.26 \\
            \midrule
            \multicolumn{5}{l}{\textit{Number of Slots}} \\
            \quad w. $m=63$           & 26.38 & 51.34 & 42.18 & 26.68 \\
            \quad w. $m=255$          & 25.31 & 46.79 & 42.82 & 34.83 \\
            \bottomrule
        \end{tabular}}
\end{wraptable}

\Cref{tab:ablations} isolates the contributions of CyFA's method and architectural choices. Removing the scalar forget gate causes the largest overall degradation: LAMBADA perplexity more than doubles from 48.17 to 115.89, and the recall average falls by 5.20 points. Replacing the input-dependent clock with one token-independent increment per head lowers recall by 2.93 points. Fixing the readout matrix to $\mathbf R=\mathbf I$ produces a further 1.46-point recall drop and raises LAMBADA perplexity to 58.75. Among the backbone components, QK normalization has a larger effect on recall than ShortConv, with drops of 3.13 and 1.55 points when they are removed. Adding the SiLU feature map commonly used in linear attention improves both perplexities but reduces the recall average from 33.25 to 32.26, so CyFA omits it. Recall is most sensitive to the number of relative-time slots: reducing $m$ from 127 to 63 lowers the recall average to 26.68, whereas increasing $m$ to 255 raises it to 34.83 and improves the other three reported metrics.
\par
\ifnum\value{WF@wrappedlines}>1
    \vspace{\dimexpr\value{WF@wrappedlines}\baselineskip-\baselineskip\relax}
\fi
\WFclear
\endgroup

\subsection{Clock Increment Analysis}

\begingroup
\setlength{\intextsep}{0pt}
\begin{wraptable}[7]{r}{0.44\linewidth}
    \vspace{-2\baselineskip}
    \centering
    \captionsetup{skip=1pt,justification=centering,singlelinecheck=false}
    \caption{Clock increments for repeated tokens in different contexts.}
    \label{tab:clock_increment_cases}
    \footnotesize
    \setlength{\tabcolsep}{3.5pt}
    \renewcommand{\arraystretch}{1.00}
    \resizebox{\linewidth}{!}{%
    \begin{tabular}{@{}llc@{}}
        \toprule
        \textbf{Head / Word} & \textbf{Context} & $\boldsymbol{\delta_t}$ \\
        \midrule
        \multirow{2}{*}{\makecell[l]{L15H3\\ \texttt{about}}}
            & ``about 45 km'' & 0.355 \\
            & ``thought about what causes'' & 0.057 \\
        \midrule
        \multirow{2}{*}{\makecell[l]{L13H2\\ \texttt{single}}}
            & Musical noun & 0.815 \\
            & Numerical modifier & 0.297 \\
        \bottomrule
    \end{tabular}}
\end{wraptable}

We examine the clock increments produced by all 96 heads of the 400M CyFA model and find that 56 heads ($58.3\%$) produce increments concentrated near 1, while the remaining 40 heads ($41.7\%$) assign substantially different increments across tokens. As an additional case study, \cref{tab:clock_increment_cases} shows two instances in which the same token receives markedly different clock increments in different contexts.
\par
\endgroup

\section{Related Work}

Linear RNNs improve fixed-size memory through increasingly selective state updates. RetNet and Mamba-2 use fixed or input-dependent scalar forget gates, whereas GLA and HGRN2 use vector-valued forget gates \citep{sun2023retnet,dao2024mamba2,yang2024gla,qin2024hgrn2}. The Delta Rule instead erases the association addressed by the incoming key before writing the new value, and Gated DeltaNet and KDA combine this key-dependent erasure with scalar and vector-valued forgetting \citep{schlag2021fastweight,yang2024gateddelta,kimiteam2025kimi}. Recent variants further introduce low-rank feedback \citep{hu2025comba}, asymmetric removal and writing \citep{peng2025rwkv7}, multiple Householder updates \citep{siems2025deltaproduct}, preconditioning \citep{tumma2026preconditioned}, momentum \citep{huang2026mdn}, decoupled erase and write controls \citep{hatamizadeh2026gdn2,li2026eda}, or \mbox{uncertainty-aware updates~\citep{bui2026kalman}}. Their transitions remain structured enough for chunk-wise training, but require richer block algebra such as compact WY and UT factorizations, DPLR products, triangular solves, or auxiliary recurrent scans \citep{yang2024deltanetparallel,kimiteam2025kimi}. CyFA follows a different direction: it reorganizes associations within a fixed state, while its absolute-clock formulation reduces both recurrent passes to scalar decay and rank-one writes.

Slot-based memories change where information is written. ABC and GSA use dense input-dependent controls over a fixed set of key and value slots, while Raven sparsely updates a selected subset \citep{peng2022abc,zhang2024gsa,afzal2026raven}. Other methods increase memory capacity through multiple routed states \citep{du2025mom}, sparsely expanded partitions or memory tables \citep{pan2025sse,cabannes2026sdm}, higher-order tensor states \citep{herranzcelotti2026runningtensor}, or an auxiliary exact KV cache \citep{cui2026hippocampus}. Log-Linear Attention and DLA instead retain multiple temporal summaries, and DART performs attention over stored Mamba-2 chunk states \citep{guo2025loglinear,wang2026dla,qian2026dart}. These methods introduce additional states, expanded storage, or a cache that grows with the number of retained summaries. CyFA keeps a fixed-size pair of key and value states. Every write enters the age-zero slot, cyclic transport organizes the state by model age, and the softmax readout remains content-based.

Among fixed-size methods, BLA is closest to CyFA's temporal organization. It reconstructs a fixed-resolution Fourier--Dirichlet window from separate key and value states, with temporal position tied affinely to token distance \citep{laborieux2026bla}. A constant-clock specialization of CyFA recovers the same temporal profile, but CyFA learns the spacing between successive writes, separates scalar forgetting from cyclic transport, and applies a learned readout over the reconstructed relative-time slots. RetNet, Selective RoPE, and Mamba-3 also use fixed or input-dependent rotations, but their rotations act on content or query--key channels \citep{sun2023retnet,movahedi2025selectiverope,lahoti2026mamba3}. CyFA applies rotation along the relative-time slot axis shared by its key and value states.

\section{Conclusion}

We introduced CyFA to improve how a fixed recurrent state organizes past
key--value associations. CyFA forms relative-time-partitioned memory by
inserting each new association into the age-zero slot and using cyclic
transport, controlled by a learned clock, to move the existing key and value
states through the relative-time slots. We derived an equivalent representation
in absolute-clock coordinates, yielding a two-pass scalar-decay formulation
that supports efficient chunk-wise training. Experiments show that CyFA
improves performance on recall-intensive tasks while maintaining competitive
language modeling and computational efficiency.

\printbibliography

\newpage
\appendix
\section{Cyclic Transport and Relative-Time Coordinates}
\label{app:temporal-addressing}
\label{app:cyclic_representation}

The main text defines CyFA through cyclic transport in relative-time slots and
then derives the absolute-clock coordinates used for efficient computation.
This section develops the mathematical structure underlying these two
representations. We first establish the Fourier construction and interpolation
properties of fractional cyclic shifts. We then examine how the learned clock
organizes stored writes by model age and prove the exact equivalence between
the relative-time and absolute-clock recurrences. Finally, we provide a
complementary temporal-address interpretation of the two-pass readout and
derive the fixed-rate specialization related to BLA.

Throughout, $m$ is the odd number of active relative-time slots,
$r,s\in\{0,\ldots,m-1\}$ index these slots, and $\lambda_t$ is the cumulative
clock defined in \cref{sec:temporal_memory}. The additional storage row used by
the implementation is discussed at the end of this section.

\subsection{Fractional Cyclic Shifts in a Real Fourier Basis}
\label{app:continuous-shifts}

We begin by making the real Fourier basis in
\cref{eq:cyfa-fractional-shift} explicit. Its columns consist of one DC
component followed by cosine--sine pairs:
\begin{align}
\boldsymbol{\Phi}_{r,0}
&= \frac{1}{\sqrt m},
&
\boldsymbol{\Phi}_{r,2j-1}
&= \sqrt{\frac{2}{m}}\cos\!\left(\frac{2\pi jr}{m}\right),
&
\boldsymbol{\Phi}_{r,2j}
&= \sqrt{\frac{2}{m}}\sin\!\left(\frac{2\pi jr}{m}\right),
\label{eq:app-real-fourier}
\end{align}
for $j=1,\ldots,(m-1)/2$. Consequently,
\begin{equation}
\boldsymbol b
= \boldsymbol{\Phi}^{\top}\boldsymbol e_0
= \left[
\frac{1}{\sqrt m},
\sqrt{\frac{2}{m}},0,\ldots,
\sqrt{\frac{2}{m}},0
\right]^{\top}.
\label{eq:app-b-explicit}
\end{equation}

\par\noindent\textbf{Proposition 1 (fractional cyclic-shift group).}
The matrix $\boldsymbol{\Phi}$ is orthogonal. For every $\tau\in\mathbb R$, $\mathcal U(\tau)$ and $\mathbf P(\tau)=\boldsymbol{\Phi}\mathcal U(\tau)\boldsymbol{\Phi}^{\top}$ are orthogonal, and
\begin{equation}
\mathbf P(\tau_1)\mathbf P(\tau_2)
= \mathbf P(\tau_1+\tau_2),
\qquad
\mathbf P(\tau)^{-1}=\mathbf P(-\tau),
\qquad
\mathbf P(\tau+m)=\mathbf P(\tau).
\label{eq:app-group-properties}
\end{equation}
Moreover, $\mathbf P(1)$ is exactly the cyclic permutation $\mathbf P$ defined in \cref{sec:temporal_memory}.

\par\noindent\textit{Proof.}
The roots-of-unity identity
\begin{equation}
\sum_{r=0}^{m-1}\exp\!\left(\frac{2\pi \mathrm i \ell r}{m}\right)
=
\begin{cases}
 m, & \ell\equiv 0 \pmod m,\\
 0, & \text{otherwise}
\end{cases}
\label{eq:app-root-sum}
\end{equation}
establishes the orthogonality of the DC, cosine, and sine columns in
Eq.~\eqref{eq:app-real-fourier}. Together with the chosen normalization, it
gives $\boldsymbol{\Phi}^{\top}\boldsymbol{\Phi}=\mathbf I$. Each block of
$\mathcal U(\tau)$ is a planar rotation. Hence,
\begin{equation}
\mathcal U(\tau)^{\top}=\mathcal U(-\tau),
\qquad
\mathcal U(\tau_1)\mathcal U(\tau_2)=\mathcal U(\tau_1+\tau_2).
\label{eq:app-rotation-properties}
\end{equation}
Conjugating these identities by $\boldsymbol{\Phi}$ yields
\cref{eq:app-group-properties}. Every frequency is an integer multiple of
$2\pi/m$, so $\mathcal U(m)=\mathbf I$, which gives the periodicity of
$\mathbf P(\tau)$.

It remains to identify the unit shift. The cyclic permutation defined in
\cref{sec:temporal_memory} acts as
$(\mathbf P\boldsymbol x)_r=\boldsymbol x_{r-1\bmod m}$. Applied to the
cosine and sine basis functions, it gives
\begin{align}
\mathbf P\cos\!\left(\frac{2\pi jr}{m}\right)
&=
\cos\!\left(\frac{2\pi j}{m}\right)
\cos\!\left(\frac{2\pi jr}{m}\right)
+
\sin\!\left(\frac{2\pi j}{m}\right)
\sin\!\left(\frac{2\pi jr}{m}\right),\\
\mathbf P\sin\!\left(\frac{2\pi jr}{m}\right)
&=
-\sin\!\left(\frac{2\pi j}{m}\right)
\cos\!\left(\frac{2\pi jr}{m}\right)
+
\cos\!\left(\frac{2\pi j}{m}\right)
\sin\!\left(\frac{2\pi jr}{m}\right).
\end{align}
Thus the action of $\mathbf P$ on each cosine--sine pair is represented by
$\operatorname{Rot}(2\pi j/m)$, while the DC component remains fixed.
Therefore $\boldsymbol{\Phi}^{\top}\mathbf P\boldsymbol{\Phi}=\mathcal U(1)$
and $\mathbf P=\mathbf P(1)$. \hfill$\square$

The group law allows successive clock increments to accumulate into a single
transport distance. Orthogonality preserves norms and inner products during
transport, while $\mathbf P(1)=\mathbf P$ connects the fractional construction
to the discrete cyclic permutation introduced in the main text.

\subsection{Transport Between Relative-Time Slots}
\label{app:cardinal-coordinates}

To characterize how fractional transport distributes stored content across the
relative-time slots, consider a state contribution initially confined to slot
$s$. After applying $\mathbf P(\tau)$, its coefficient at slot $r$ is
$\boldsymbol e_r^\top\mathbf P(\tau)\boldsymbol e_s$. The Fourier
representation of the integer slots follows from
Eqs.~\eqref{eq:app-real-fourier}--\eqref{eq:app-b-explicit}:
\begin{equation}
\boldsymbol{\Phi}^{\top}\boldsymbol e_r
= \mathcal U(r)\boldsymbol b,
\qquad r=0,\ldots,m-1.
\label{eq:app-frame-orbit}
\end{equation}
This identity expresses each relative-time slot in the real Fourier basis.

\par\noindent\textbf{Proposition 2 (fractional cyclic-shift kernel).}
For any $r,s\in\{0,\ldots,m-1\}$ and $\tau\in\mathbb R$,
\begin{align}
\boldsymbol e_r^{\top}\mathbf P(\tau)\boldsymbol e_s
&=
\boldsymbol b^{\top}\mathcal U(\tau+s-r)\boldsymbol b
\label{eq:app-kernel-inner-product}\\
&=
\frac{1}{m}
\left[
1+2\sum_{j=1}^{(m-1)/2}
\cos\!\left(\frac{2\pi j(\tau+s-r)}{m}\right)
\right]
\label{eq:app-kernel-sum}\\
&=
\frac{\sin\!\big(\pi(\tau+s-r)\big)}
{m\sin\!\big(\pi(\tau+s-r)/m\big)},
\label{eq:app-kernel-closed}
\end{align}
where the final ratio is interpreted by continuity when its denominator vanishes.

\par\noindent\textit{Proof.}
Using \cref{eq:app-frame-orbit}, the orthogonality of $\mathcal U$, and its
composition law gives
\begin{align}
\boldsymbol e_r^{\top}\mathbf P(\tau)\boldsymbol e_s
&=
(\boldsymbol{\Phi}^{\top}\boldsymbol e_r)^{\top}
\mathcal U(\tau)
(\boldsymbol{\Phi}^{\top}\boldsymbol e_s)\\
&=
(\mathcal U(r)\boldsymbol b)^{\top}
\mathcal U(\tau)
\mathcal U(s)\boldsymbol b
=
\boldsymbol b^{\top}\mathcal U(\tau+s-r)\boldsymbol b.
\end{align}
Substituting \cref{eq:app-b-explicit} gives
\cref{eq:app-kernel-sum}. Writing the cosine sum as a symmetric geometric
series yields
\begin{equation}
\frac{1}{m}
\sum_{j=-(m-1)/2}^{(m-1)/2}
\exp\!\left(\frac{2\pi \mathrm i j x}{m}\right)
=
\frac{\sin(\pi x)}{m\sin(\pi x/m)},
\end{equation}
and setting $x=\tau+s-r$ gives
\cref{eq:app-kernel-closed}. \hfill$\square$

The resulting kernel gives the coefficient with which content initially placed
in slot $s$ is transported to slot $r$.

\paragraph{Cardinality.}
For integer $a$, Eq.~\eqref{eq:app-kernel-closed} is one when $r\equiv s+a\pmod m$ and zero at every other integer slot. Therefore
\begin{equation}
\mathbf P(a)\boldsymbol e_s=\boldsymbol e_{s+a\bmod m}.
\label{eq:app-cardinality}
\end{equation}
Integer transport thus moves the complete contribution from one relative-time
slot to another without distributing it across the remaining slots.

\paragraph{Partition of unity and energy preservation.}
The DC component is fixed by every $\mathcal U(\tau)$, while all non-DC components sum to zero over the integer grid. Hence
\begin{equation}
\boldsymbol 1^{\top}\mathbf P(\tau)\boldsymbol e_s=1.
\label{eq:app-partition-unity}
\end{equation}
Because $\mathbf P(\tau)$ is orthogonal,
\begin{equation}
\left\|\mathbf P(\tau)\boldsymbol e_s\right\|_2=1,
\qquad
\big(\mathbf P(\tau)\boldsymbol e_s\big)^{\top}
\big(\mathbf P(\tau)\boldsymbol e_{s'}\big)
=\delta_{s,s'}.
\label{eq:app-profile-orthogonality}
\end{equation}
For every fractional shift, the $m$ transported basis profiles therefore remain
a complete orthonormal coordinate system. A fractional cyclic shift
continuously interpolates between the integer relative-time slots while
preserving this geometry. The interpolation coefficients can have signed
sidelobes and should not be interpreted as probabilities. Probabilistic
normalization enters only through the softmax readout in
\cref{eq:cyfa-physical-readout}.

\subsection{Evolution of Stored Writes}
\label{app:relative-memory}
\label{app:partition-geometry}

We now follow each write from its insertion at the age-zero slot through the
subsequent state transitions.

\par\noindent\textbf{Proposition 3 (unrolled relative-time states).}
Starting from zero state, the recurrence in
\cref{eq:cyfa-physical-recurrence} gives
\begin{align}
\mathbf K_t
&=
\sum_{i\le t}
\beta_i
\left(\prod_{j=i+1}^{t}\alpha_j\right)
\mathbf P(\lambda_t-\lambda_i)\boldsymbol e_0\mathbf k_i^{\top},
\label{eq:app-unrolled-K}\\
\mathbf V_t
&=
\sum_{i\le t}
\beta_i
\left(\prod_{j=i+1}^{t}\alpha_j\right)
\mathbf P(\lambda_t-\lambda_i)\boldsymbol e_0\mathbf v_i^{\top}.
\label{eq:app-unrolled-V}
\end{align}
Consequently, row $r$ of the relative-time key state is
\begin{equation}
\boldsymbol e_r^{\top}\mathbf K_t
=
\sum_{i\le t}
\beta_i
\left(\prod_{j=i+1}^{t}\alpha_j\right)
\big[\mathbf P(\lambda_t-\lambda_i)\boldsymbol e_0\big]_r
\mathbf k_i^{\top}.
\label{eq:app-row-decomposition}
\end{equation}

\par\noindent\textit{Proof.}
The statement is immediate at $t=0$. Assuming
Eq.~\eqref{eq:app-unrolled-K} at $t-1$,
\cref{eq:cyfa-physical-recurrence} and the group law give
\begin{align}
\mathbf K_t
&=
\alpha_t\mathbf P(\delta_t)
\sum_{i\le t-1}
\beta_i
\left(\prod_{j=i+1}^{t-1}\alpha_j\right)
\mathbf P(\lambda_{t-1}-\lambda_i)\boldsymbol e_0\mathbf k_i^{\top}
+
\beta_t\boldsymbol e_0\mathbf k_t^{\top}\\
&=
\sum_{i\le t-1}
\beta_i
\left(\prod_{j=i+1}^{t}\alpha_j\right)
\mathbf P(\delta_t+\lambda_{t-1}-\lambda_i)\boldsymbol e_0\mathbf k_i^{\top}
+
\beta_t\boldsymbol e_0\mathbf k_t^{\top}\\
&=
\sum_{i\le t}
\beta_i
\left(\prod_{j=i+1}^{t}\alpha_j\right)
\mathbf P(\lambda_t-\lambda_i)\boldsymbol e_0\mathbf k_i^{\top}.
\end{align}
The value-state identity is identical. Left-multiplying by
$\boldsymbol e_r^{\top}$ yields Eq.~\eqref{eq:app-row-decomposition}.
\hfill$\square$

The factor $\mathbf P(\lambda_t-\lambda_i)\boldsymbol e_0$ composes every
fractional cyclic shift applied after write $i$ entered at the age-zero slot.
It is therefore the relative-time profile of write $i$ at step $t$. In
\cref{eq:app-unrolled-K}, the write strength and subsequent scalar forget gates
determine the magnitude of each state contribution, while its relative-time
profile determines how that contribution is distributed across the slots.

\par\noindent\textbf{Proposition 4 (relative-time geometry under cyclic transport).}
For every previously written token $i\leq t$,
\begin{equation}
\mathbf P(\lambda_{t+1}-\lambda_i)\boldsymbol e_0
=
\mathbf P(\delta_{t+1})
\mathbf P(\lambda_t-\lambda_i)\boldsymbol e_0.
\label{eq:app-coherent-transport}
\end{equation}
Moreover, for any two writes $i,j\leq t$, the inner product between their
relative-time profiles is independent of the current step:
\begin{align}
&\big(\mathbf P(\lambda_t-\lambda_i)\boldsymbol e_0\big)^{\top}
 \big(\mathbf P(\lambda_t-\lambda_j)\boldsymbol e_0\big)
\label{eq:app-profile-overlap-start}\\
&\qquad=
\boldsymbol e_0^{\top}\mathbf P(\lambda_i-\lambda_j)\boldsymbol e_0
=
\frac{
\sin\!\big(\pi(\lambda_i-\lambda_j)\big)
}{
m\sin\!\big(\pi(\lambda_i-\lambda_j)/m\big)
}.
\label{eq:app-profile-overlap}
\end{align}

\par\noindent\textit{Proof.}
Equation~\eqref{eq:app-coherent-transport} follows from
$\lambda_{t+1}=\lambda_t+\delta_{t+1}$ and the group law in Proposition~1. For
the inner product, orthogonality and the same group law give
\begin{align}
&\big(\mathbf P(\lambda_t-\lambda_i)\boldsymbol e_0\big)^{\top}
 \big(\mathbf P(\lambda_t-\lambda_j)\boldsymbol e_0\big)\\
&\qquad=
\boldsymbol e_0^{\top}
\mathbf P(\lambda_i-\lambda_t)
\mathbf P(\lambda_t-\lambda_j)
\boldsymbol e_0
=
\boldsymbol e_0^{\top}\mathbf P(\lambda_i-\lambda_j)\boldsymbol e_0.
\end{align}
The closed form is Proposition~2 with $r=s=0$. \hfill$\square$

Every clock step therefore applies the same orthogonal cyclic shift to all
existing writes. This common transport preserves their pairwise profile
geometry while advancing them relative to the age-zero slot where the next
write is inserted.

\par\noindent\textbf{Proposition 5 (order-preserving adaptive spacing).}
For any $i<j\leq t$,
\begin{equation}
\big(\lambda_t-\lambda_i\big)-\big(\lambda_t-\lambda_j\big)
=
\lambda_j-\lambda_i
=
\sum_{u=i+1}^{j}\delta_u
\in (0,j-i).
\label{eq:app-order-preserving-clock}
\end{equation}
Hence the learned clock preserves chronological order: at every later readout,
write $i$ has a strictly larger model age than write $j$. At the same time, the
separation assigned to the interval $i+1,\ldots,j$ is data dependent.

\par\noindent\textit{Proof.}
The first equality cancels the common current clock $\lambda_t$. The second
follows by telescoping $\lambda_u=\lambda_{u-1}+\delta_u$. Since every
$\delta_u\in(0,1)$, the sum is strictly between $0$ and $j-i$.
\hfill$\square$

In particular,
\begin{equation}
\lambda_t-\lambda_i
=
\sum_{u=i+1}^{t}\delta_u,
\qquad
\lambda_t-\lambda_i<m
\iff
\sum_{u=i+1}^{t}\delta_u<m.
\label{eq:app-data-dependent-occupancy}
\end{equation}
The number of token positions traversed before a write completes one cycle is
therefore determined by the accumulated clock increments rather than by a fixed
token window. Small increments compress more consecutive tokens into one unit
of model age, while larger increments separate them more strongly.

Because the relative-time coordinates are cyclic, model ages that differ by an
integer number of complete cycles are mapped to the same cyclic position. Their
state contributions retain their respective write strengths and accumulated
scalar forget-gate factors from Proposition~3, which determine their remaining
magnitudes across repeated cycles.

\par\noindent\textbf{Corollary 1 (overlap of transported state contributions).}
After factoring out the scalar write-and-forget coefficients in Proposition~3,
the Frobenius inner product between the key-state contributions of writes $i$
and $j$ is
\begin{align}
&\left\langle
\mathbf P(\lambda_t-\lambda_i)\boldsymbol e_0\boldsymbol k_i^{\top},
\mathbf P(\lambda_t-\lambda_j)\boldsymbol e_0\boldsymbol k_j^{\top}
\right\rangle_{\mathrm F}
\label{eq:app-factorized-overlap-start}\\
&\qquad=
(\boldsymbol k_i^{\top}\boldsymbol k_j)
\frac{
\sin\!\big(\pi(\lambda_i-\lambda_j)\big)
}{
m\sin\!\big(\pi(\lambda_i-\lambda_j)/m\big)
}.
\label{eq:app-factorized-overlap}
\end{align}
The value-state contributions satisfy the same identity with
$\boldsymbol k_i^{\top}\boldsymbol k_j$ replaced by
$\boldsymbol v_i^{\top}\boldsymbol v_j$.

\par\noindent\textit{Proof.}
For vectors $\boldsymbol a,\boldsymbol c$ and $\boldsymbol x,\boldsymbol y$,
$\langle \boldsymbol a\boldsymbol x^{\top},
\boldsymbol c\boldsymbol y^{\top}\rangle_{\mathrm F}
=(\boldsymbol a^{\top}\boldsymbol c)(\boldsymbol x^{\top}\boldsymbol y)$.
Applying this identity and Proposition~4 gives
Eq.~\eqref{eq:app-factorized-overlap}. The omitted amplitudes simply multiply
the two sides. \hfill$\square$

This factorization separates content similarity from relative-time-profile
overlap. The learned clock determines the second factor when the writes enter
the relative-time state, and subsequent common shifts preserve it during
transport.

\subsection{Absolute-Clock Coordinates as an Exact Change of Coordinates}
\label{app:absolute-coordinates}

The three key-state representations used in the main text are
$\mathbf K_t$ in relative-time slots,
$\widehat{\mathbf K}_t$ in Fourier coordinates, and
$\overline{\mathbf K}_t$ in absolute-clock coordinates, with
\begin{equation}
\widehat{\mathbf K}_t=\boldsymbol{\Phi}^{\top}\mathbf K_t,
\qquad
\overline{\mathbf K}_t=\mathcal U(-\lambda_t)\widehat{\mathbf K}_t,
\label{eq:app-three-K}
\end{equation}
and identical definitions for $\mathbf V_t$. These coordinates recover the two
parts of the relative-time update separately:
\begin{align}
\boldsymbol\Phi\mathcal U(\lambda_t)\overline{\mathbf K}_{t-1}
&=
\boldsymbol\Phi\mathcal U(\lambda_t-\lambda_{t-1})
\boldsymbol\Phi^\top\mathbf K_{t-1}
=
\mathbf P(\delta_t)\mathbf K_{t-1},
\label{eq:app-reconstruct-shifted-state}\\
\boldsymbol\Phi\mathcal U(\lambda_t)\mathcal U(-\lambda_t)\boldsymbol b
&=
\boldsymbol\Phi\boldsymbol b
=
\boldsymbol e_0.
\label{eq:app-reconstruct-entry}
\end{align}
The temporal write vector is the age-zero insertion expressed in
absolute-clock coordinates. Reconstruction at the current clock maps the
previous relative-time state to its shifted version and the temporal write
vector to $\boldsymbol e_0$. The scalar recurrence in
\cref{eq:cyfa-absolute-recurrence} thus implements exactly the
transport-and-insert update in \cref{eq:cyfa-physical-recurrence}.

The unrolled states show how this equivalence persists over the complete
history. Applying $\boldsymbol{\Phi}^{\top}$ to
Eq.~\eqref{eq:app-unrolled-K} gives
\begin{equation}
\widehat{\mathbf K}_t
=
\sum_{i\le t}
\beta_i
\left(\prod_{j=i+1}^{t}\alpha_j\right)
\mathcal U(\lambda_t-\lambda_i)\boldsymbol b\mathbf k_i^{\top}.
\label{eq:app-relative-fourier-expanded}
\end{equation}
Multiplying by $\mathcal U(-\lambda_t)$ then yields
\begin{equation}
\overline{\mathbf K}_t
=
\sum_{i\le t}
\beta_i
\left(\prod_{j=i+1}^{t}\alpha_j\right)
\mathcal U(-\lambda_i)\boldsymbol b\mathbf k_i^{\top},
\label{eq:app-absolute-expanded}
\end{equation}
The value state follows identically with $\mathbf v_i^\top$ in place of
$\mathbf k_i^\top$. In absolute-clock coordinates, the direction representing
each earlier write remains fixed. Its relative-time profile is recovered by the
current reconstruction map. Equation~\eqref{eq:app-absolute-expanded}
also directly implies the scalar-decay recurrence in
\cref{eq:cyfa-absolute-recurrence}.

The complete relative-time state is recovered exactly:
\begin{align}
\boldsymbol{\Phi}\mathcal U(\lambda_t)\overline{\mathbf K}_t
&=
\sum_{i\le t}
\beta_i
\left(\prod_{j=i+1}^{t}\alpha_j\right)
\boldsymbol{\Phi}\mathcal U(\lambda_t-\lambda_i)\boldsymbol b\mathbf k_i^{\top}\\
&=
\sum_{i\le t}
\beta_i
\left(\prod_{j=i+1}^{t}\alpha_j\right)
\mathbf P(\lambda_t-\lambda_i)\boldsymbol e_0\mathbf k_i^{\top}
=
\mathbf K_t.
\label{eq:app-reconstruct-relative}
\end{align}
The same reconstruction holds for $\mathbf V_t$. Substitution into
\cref{eq:cyfa-physical-readout} gives
\cref{eq:cyfa-absolute-readout}, so both the update and the layer output are
unchanged by the coordinate transformation.

\par\noindent\textbf{Proposition 6 (clock-origin invariance).}
Let every clock value be shifted by an arbitrary constant $c\in\mathbb R$, so that $\lambda_t'=\lambda_t+c$. If the absolute-clock states are formed using the shifted temporal write vectors, then
\begin{equation}
\overline{\mathbf K}_t'=\mathcal U(-c)\overline{\mathbf K}_t,
\qquad
\overline{\mathbf V}_t'=\mathcal U(-c)\overline{\mathbf V}_t,
\label{eq:app-gauge-state}
\end{equation}
and the output in \cref{eq:cyfa-absolute-readout} is unchanged.

\par\noindent\textit{Proof.}
Equation~\eqref{eq:app-gauge-state} follows term by term from
$\mathcal U(-\lambda_i-c)=\mathcal U(-c)\mathcal U(-\lambda_i)$. At readout,
\begin{equation}
\boldsymbol{\Phi}\mathcal U(\lambda_t+c)\overline{\mathbf K}_t'
=
\boldsymbol{\Phi}\mathcal U(\lambda_t+c)\mathcal U(-c)\overline{\mathbf K}_t
=
\boldsymbol{\Phi}\mathcal U(\lambda_t)\overline{\mathbf K}_t,
\end{equation}
and the same identity holds for the value state. Substitution into \cref{eq:cyfa-physical-readout}, or equivalently \cref{eq:cyfa-absolute-readout}, leaves $\mathbf o_t$ unchanged. \hfill$\square$

The clock origin is therefore arbitrary. Although \cref{eq:cyfa-absolute-recurrence} uses clock values computationally, the represented memory and the layer output depend only on model ages $\lambda_t-\lambda_i$.

\subsection{Temporal Addresses and the Two-Pass Readout}
\label{app:temporal-address-view}

In the recurrence, $\mathcal U(-\lambda_i)\boldsymbol b$ is the temporal write
vector that inserts write $i$ into the absolute-clock state. The same vector
also admits a complementary address-matching interpretation. Viewed in this
way, $\mathcal U(-\lambda)\boldsymbol b$ forms a continuous family of temporal
addresses, while $\mathbf k_i$ and $\mathbf v_i$ carry the key and value
content.

The similarity between two such addresses is
\begin{align}
\big(\mathcal U(-\lambda_i)\boldsymbol b\big)^{\top}
\big(\mathcal U(-\lambda_j)\boldsymbol b\big)
&=
\boldsymbol b^{\top}\mathcal U(\lambda_i-\lambda_j)\boldsymbol b\\
&=
\frac{\sin\!\big(\pi(\lambda_i-\lambda_j)\big)}
{m\sin\!\big(\pi(\lambda_i-\lambda_j)/m\big)}.
\label{eq:app-address-similarity}
\end{align}
Addresses whose clock values differ by a nonzero integer within a cycle are
orthogonal. Fractional separations produce the corresponding cardinal
interpolation similarity.

The first pass in \cref{eq:cyfa-two-pass} gives
\begin{equation}
\mathbf o_t'
=
\overline{\mathbf K}_t\mathbf q_t
=
\sum_{i\le t}
\beta_i
\left(\prod_{j=i+1}^{t}\alpha_j\right)
(\mathbf k_i^{\top}\mathbf q_t)
\mathcal U(-\lambda_i)\boldsymbol b.
\label{eq:app-first-pass-addresses}
\end{equation}
The key--query match $\mathbf k_i^{\top}\mathbf q_t$ therefore scales the
temporal address of write $i$. The first pass does not yet retrieve a value. It
accumulates key evidence in the temporal-address space.

The $r$-th canonical relative-time slot at token $t$ corresponds, in absolute-clock coordinates, to the query address
\begin{equation}
\boldsymbol e_r^{\top}\boldsymbol{\Phi}\mathcal U(\lambda_t)
=
\big(\mathcal U(r-\lambda_t)\boldsymbol b\big)^{\top}.
\label{eq:app-relative-address-query}
\end{equation}
Its match to the address of write $i$ is
\begin{align}
\big(\mathcal U(r-\lambda_t)\boldsymbol b\big)^{\top}
\mathcal U(-\lambda_i)\boldsymbol b
&=
\boldsymbol b^{\top}\mathcal U(\lambda_t-\lambda_i-r)\boldsymbol b\\
&=
\big[\mathbf P(\lambda_t-\lambda_i)\boldsymbol e_0\big]_r.
\label{eq:app-relative-address-match}
\end{align}
Combining Eqs.~\eqref{eq:app-first-pass-addresses} and \eqref{eq:app-relative-address-match}, the contribution of write $i$ to relative-time slot $r$ before the learned readout is
\begin{equation}
\beta_i
\left(\prod_{j=i+1}^{t}\alpha_j\right)
(\mathbf k_i^{\top}\mathbf q_t)
\big[\mathbf P(\lambda_t-\lambda_i)\boldsymbol e_0\big]_r.
\label{eq:app-write-logit-contribution}
\end{equation}
The address inner product in
\cref{eq:app-relative-address-match} is exactly the transport coefficient
derived in Proposition~2. The same cardinal interpolation kernel therefore
governs cyclic transport in relative-time coordinates and address matching in
absolute-clock coordinates.

The learned readout matrix $\mathbf R$ forms task-adaptive linear combinations
of the canonical address queries. Row $s$ of the readout matrix satisfies
\begin{equation}
\boldsymbol e_s^{\top}\mathbf R\boldsymbol{\Phi}\mathcal U(\lambda_t)
=
\sum_{r=0}^{m-1}
\mathbf R_{s,r}
\big(\mathcal U(r-\lambda_t)\boldsymbol b\big)^{\top}.
\label{eq:app-learned-address-query}
\end{equation}
Thus $\mathbf R$ learns a bank of temporal-address queries anchored to model
age. Because the current clock value is applied before $\mathbf R$, each row of
the readout matrix retains the same relative-time meaning for every clock
value.

The transpose transformation in the middle line of
\cref{eq:cyfa-two-pass} has the matching address interpretation. For any
slot-weight vector $\mathbf w\in\mathbb R^m$,
\begin{equation}
\mathcal U(-\lambda_t)\boldsymbol{\Phi}^{\top}\mathbf R^{\top}\mathbf w
=
\sum_{r=0}^{m-1}
[\mathbf R^{\top}\mathbf w]_r
\mathcal U(r-\lambda_t)\boldsymbol b.
\label{eq:app-address-mixture}
\end{equation}
After setting $\mathbf w$ to the softmax weights in
\cref{eq:cyfa-two-pass}, the second pass queries the value state with this
mixture of current relative-time addresses. Since
$\overline{\mathbf V}_t$ associates every $\mathbf v_i$ with the same temporal
address $\mathcal U(-\lambda_i)\boldsymbol b$, the inner products in
\cref{eq:app-relative-address-match} recover exactly the corresponding
relative-time value aggregation.

\subsection{Fixed-Rate Specialization and Relation to BLA}
\label{app:bla}

If the learned clock is constrained to a constant increment
$\delta_u\equiv 1/\rho$, then
$\lambda_t-\lambda_i=(t-i)/\rho$. Writing $T=\rho m$, Proposition~2 gives
\begin{equation}
\big[
\mathbf P\!\left((t-i)/\rho\right)\boldsymbol e_0
\big]_r
=
\frac{1}{m}
\left[
1+2\sum_{h=1}^{(m-1)/2}
\cos\!\left(
\frac{2\pi h(t-i-\rho r)}{T}
\right)
\right].
\label{eq:app-constant-rate-profile}
\end{equation}
This is the fixed-resolution Fourier--Dirichlet profile used by Blurry Window
Attention (BLA) \citep{laborieux2026bla}. Under this constraint, every
$\rho$ token steps advance one relative-time slot and one cycle spans $T$ token
steps. CyFA retains the same Fourier--Dirichlet profile family without imposing
an affine relation between token distance and model age. The cumulative
increments in \cref{eq:app-data-dependent-occupancy} instead determine the
transport distance.

Under the fixed-rate specialization, CyFA and BLA share the same relative-time
profile. Starting from this profile, the two methods construct their memory
updates differently. With state decay, BLA uses the profile-coupled GSA
recurrence
\begin{equation}
\mathbf K_t=
\operatorname{Diag}(\boldsymbol 1-\boldsymbol\phi_t)\mathbf K_{t-1}
+\boldsymbol\phi_t\mathbf k_t^\top,
\end{equation}
with an analogous update for $\mathbf V_t$. Its position within the cycle is
affine in the token index with period $T$. In its efficient form, BLA keeps the
cumulative state in absolute coordinates and uses joint row-permutation
invariance of key--value softmax. CyFA combines the profile family with a
data-dependent monotone clock, a separate scalar forget gate, and the learned
readout matrix $\mathbf R$. Before the readout matrix and softmax are applied,
the current clock reconstructs the relative-time profiles for the current
token.

\subsection{Active Coordinates and Padded Storage}
\label{app:padded-storage}

All derivations above operate on the $m=127$ active relative-time slots. The
implementation allocates $m^{\ast}=128$ storage rows for hardware convenience.
The additional padding row lies outside the Fourier basis, is masked during
writing and readout, and remains zero. The implemented operators therefore act
as the derived $m$-dimensional maps on the active coordinates and as zero on the
padding row. This additional storage row changes neither the recurrence nor any
of the coordinate identities above.

\section{Hardware-Efficient Chunk-Wise Parallelism}
\label{app:chunkwise_parallel}

\paragraph{CyFA states in the ScalarGatedLA interface.}
Across chunks, CyFA carries the absolute-clock states
$\overline{\mathbf K}$ and $\overline{\mathbf V}$. In the key pass, the
recurrent state of the first $\operatorname{ScalarGatedLA}$ call is
\begin{equation}
\mathbf S_t^{(K)}
=
\alpha_t\mathbf S_{t-1}^{(K)}
+
\mathbf k_t
\big(\beta_t\mathcal U(-\lambda_t)\boldsymbol b\big)^\top
=
\overline{\mathbf K}_t^\top,
\label{eq:app-first-interface-state}
\end{equation}
Its output is
$(\mathbf S_t^{(K)})^\top\mathbf q_t
=\overline{\mathbf K}_t\mathbf q_t=\mathbf o_t'$. In the value pass, the
recurrent state of the second call is
\begin{equation}
\mathbf S_t^{(V)}
=
\alpha_t\mathbf S_{t-1}^{(V)}
+
\beta_t\mathcal U(-\lambda_t)\boldsymbol b\mathbf v_t^\top
=
\overline{\mathbf V}_t,
\label{eq:app-second-interface-state}
\end{equation}
Its output is $\overline{\mathbf V}_t^\top\mathbf o_t''$, exactly as in
\cref{eq:cyfa-absolute-readout}. The temporal write vector therefore occupies
the value position in the key pass and the key position in the value pass.

\paragraph{Scalar-decay chunk-wise form.}
We follow the chunk notation of
\cref{sec:background-linear-attention}. The state after $i$ chunks is
$\mathbf{S}_{[i]}=\mathbf{S}_{iC}$, and
$\mathbf{Q}_{[i+1]},\mathbf{K}_{[i+1]},\mathbf{V}_{[i+1]}$ contain the next
$C$ tokens. Consider the $\operatorname{ScalarGatedLA}$ recurrence in
\cref{tab:memory-recurrences},
$\mathbf{S}_t=\alpha_t\mathbf{S}_{t-1}+\boldsymbol{k}_t\boldsymbol{v}_t^\top$.
For chunk $i+1$, define
\begin{equation}
    \gamma_{[i+1]}^{r}
    =
    \prod_{j=1}^{r}\alpha_{iC+j},
    \qquad
    \bigl(\boldsymbol{\Gamma}_{[i+1]}\bigr)_{rs}
    =
    \begin{cases}
        \gamma_{[i+1]}^{r}/\gamma_{[i+1]}^{s}, & r\ge s,\\
        0, & r<s,
    \end{cases}
    \label{eq:app-chunk-decay}
\end{equation}
for $r,s=1,\ldots,C$. The matrix $\boldsymbol{\Gamma}_{[i+1]}$ is the
decay-aware counterpart of the causal mask $\mathbf{M}$ in
\cref{sec:background-linear-attention}. For any block
$\mathbf{X}_{[i+1]}\in\mathbb{R}^{C\times d}$, define
\begin{equation}
\begin{aligned}
    \bigl(\overleftarrow{\mathbf{X}}_{[i+1]}\bigr)_{r:}
    &=
    \gamma_{[i+1]}^{r}\bigl(\mathbf{X}_{[i+1]}\bigr)_{r:},\\
    \bigl(\overrightarrow{\mathbf{X}}_{[i+1]}\bigr)_{r:}
    &=
    \frac{\gamma_{[i+1]}^{C}}{\gamma_{[i+1]}^{r}}
    \bigl(\mathbf{X}_{[i+1]}\bigr)_{r:}.
\end{aligned}
\label{eq:app-chunk-rescale}
\end{equation}
The scalar recurrence then admits the exact chunk-wise form
\begin{equation}
\begin{aligned}
    \mathbf{S}_{[i+1]}
    &=
    \gamma_{[i+1]}^{C}\mathbf{S}_{[i]}
    +
    \mathbf{K}_{[i+1]}^\top
    \overrightarrow{\mathbf{V}}_{[i+1]},\\
    \mathbf{O}_{[i+1]}
    &=
    \overleftarrow{\mathbf{Q}}_{[i+1]}\mathbf{S}_{[i]}
    +
    \left(
        \mathbf{Q}_{[i+1]}\mathbf{K}_{[i+1]}^\top
        \odot\boldsymbol{\Gamma}_{[i+1]}
    \right)\mathbf{V}_{[i+1]}.
\end{aligned}
\label{eq:app-scalar-chunk}
\end{equation}
The first output term retrieves from the state carried across preceding chunks.
The second computes all causal interactions within the current chunk in
parallel. In practice, the decay factors are obtained from chunk-local
cumulative sums of $\log\alpha_t$.

\paragraph{CyFA chunk-wise form.}
For the absolute-clock recurrence in
\cref{eq:cyfa-absolute-recurrence}, stack the temporal write vectors of chunk
$i+1$ into the temporal write matrix
$\mathbf{A}_{[i+1]}\in\mathbb{R}^{C\times m}$, whose $r$-th row is
\begin{equation}
    \bigl(\mathbf{A}_{[i+1]}\bigr)_{r:}
    =
    \left(
        \beta_{iC+r}\mathcal{U}(-\lambda_{iC+r})\boldsymbol{b}
    \right)^\top.
    \label{eq:app-temporal-writes}
\end{equation}
We extend the bracket notation to the absolute-clock states,
$\overline{\mathbf{K}}_{[i]}=\overline{\mathbf{K}}_{iC}$ and
$\overline{\mathbf{V}}_{[i]}=\overline{\mathbf{V}}_{iC}$.
Applying \cref{eq:app-scalar-chunk} to the key pass in
\cref{eq:cyfa-two-pass} gives
\begin{equation}
\begin{aligned}
    \overline{\mathbf{K}}_{[i+1]}
    &=
    \gamma_{[i+1]}^{C}\overline{\mathbf{K}}_{[i]}
    +
    \overrightarrow{\mathbf{A}}_{[i+1]}^\top
    \mathbf{K}_{[i+1]},\\
    \mathbf{O}'_{[i+1]}
    &=
    \overleftarrow{\mathbf{Q}}_{[i+1]}
    \overline{\mathbf{K}}_{[i]}^\top
    +
    \left(
        \mathbf{Q}_{[i+1]}\mathbf{K}_{[i+1]}^\top
        \odot\boldsymbol{\Gamma}_{[i+1]}
    \right)\mathbf{A}_{[i+1]}.
\end{aligned}
\label{eq:app-cyfa-key-chunk}
\end{equation}
The middle line of \cref{eq:cyfa-two-pass} is then applied independently to
each row of $\mathbf{O}'_{[i+1]}$, producing
$\mathbf{O}''_{[i+1]}$. Applying the same chunk-wise form to the value pass
gives
\begin{equation}
\begin{aligned}
    \overline{\mathbf{V}}_{[i+1]}
    &=
    \gamma_{[i+1]}^{C}\overline{\mathbf{V}}_{[i]}
    +
    \mathbf{A}_{[i+1]}^\top
    \overrightarrow{\mathbf{V}}_{[i+1]},\\
    \mathbf{O}_{[i+1]}
    &=
    \overleftarrow{\mathbf{O}''}_{[i+1]}
    \overline{\mathbf{V}}_{[i]}
    +
    \left(
        \mathbf{O}''_{[i+1]}\mathbf{A}_{[i+1]}^\top
        \odot\boldsymbol{\Gamma}_{[i+1]}
    \right)\mathbf{V}_{[i+1]}.
\end{aligned}
\label{eq:app-cyfa-value-chunk}
\end{equation}
Both passes use the same scalar forget gate $\alpha_t$ and therefore share the
factors $\{\gamma_{[i+1]}^r\}_{r=1}^{C}$ and the mask
$\boldsymbol{\Gamma}_{[i+1]}$. The chunk-boundary carry represents the
cumulative cyclic transport and scalar forgetting applied to the state from
preceding chunks. At the next chunk boundary,
\begin{equation}
\boldsymbol\Phi\mathcal U(\lambda_{(i+1)C})
\left(\gamma_{[i+1]}^{C}\overline{\mathbf K}_{[i]}\right)
=
\gamma_{[i+1]}^{C}
\mathbf P(\lambda_{(i+1)C}-\lambda_{iC})\mathbf K_{iC}.
\label{eq:app-chunk-transport}
\end{equation}
The new-write terms recover their respective cumulative shifts through the
same coordinate identity. After reconstruction in relative-time coordinates,
the carried state has advanced by the clock distance accumulated across the
chunk. The scalar boundary recurrence accounts for this transport without
explicitly applying the corresponding shift matrix.

\section{Experimental Details}
\label{app:experimental_details}

\subsection{Architecture and State Matching}
\label{app:architecture_state_matching}

\paragraph{Backbone and mixer configurations.}
All models pretrained for our comparison use a 24-layer pre-norm LLaMA-style
backbone. Each layer contains a sequence mixer followed by a SwiGLU
feed-forward block.

To isolate the recurrent update in the comparison with BLA, we place its
decayed recurrence within the same surrounding mixer design used by CyFA. Both
models therefore use ShortConv, RMSNorm-based QK normalization, head-wise
RMSNorm, and the same low-rank sigmoid output gate. Only the core recurrence is
changed. Because no official BLA implementation was publicly available at the
time of our experiments, we implement the recurrence described by
\citet{laborieux2026bla} in its GSA-equivalent form. The Dirichlet interpolation
profile $\boldsymbol\phi_t$ provides the slot-wise write weights, while
$\boldsymbol 1-\boldsymbol\phi_t$ controls the decay of the previous state.
This parameterization allows us to use the Triton GSA chunk operator from
\texttt{flash-linear-attention} \citep{zhang2024gsa,yang2024fla}.

Mamba-2 and Mamba-3 each use one Mamba mixer per backbone layer
\citep{dao2024mamba2,lahoti2026mamba3}, with an expansion factor of 2,
$d_{\mathrm{state}}=128$, and $d_{\mathrm{head}}=64$. For Mamba-3, we use the
rank-$4$ MIMO variant.

\paragraph{State-size accounting.}
For all state-size comparisons, we count the scalar entries in the main
recurrent state of each layer and omit the ShortConv cache.

SWA, GSA, and Raven store key and value tensors with $m$ slots per head, giving state shapes $H\times m\times d_k$ and $H\times m\times d_v$. At 400M, we use $H=4$, $m=128$, and $d_k=d_v=256$, for a total of
\begin{equation}
    Hm(d_k+d_v)=4\times128\times(256+256)=262{,}144
\label{eq:app-state-size-slot-memory}
\end{equation}
scalar entries per layer.

CyFA and BLA use $m=127$ active slots and allocate $m^\ast=m+1=128$ storage
rows for hardware-efficient computation. At $H=4$ and $d_k=d_v=256$, their
state size is
\begin{equation}
    Hm^\ast(d_k+d_v)=4\times128\times(256+256)=262{,}144.
\label{eq:app-state-size-cyfa-bla}
\end{equation}

GDN and KDA maintain one $d_v\times d_k$ state matrix per head. With $H=4$ and $d_k=d_v=256$, their state size is
\begin{equation}
    Hd_kd_v=4\times256\times256=262{,}144.
\label{eq:app-state-size-matrix}
\end{equation}

Mamba-2 and Mamba-3 maintain a recurrent state of shape
$H\times d_{\mathrm{head}}\times d_{\mathrm{state}}$. At $d=1024$, an
expansion factor of 2 and $d_{\mathrm{head}}=64$ give
\begin{equation}
    H=\frac{2d}{d_{\mathrm{head}}}=32.
\label{eq:app-mamba-head-count}
\end{equation}
Together with $d_{\mathrm{state}}=128$, this yields
\begin{equation}
    Hd_{\mathrm{head}}d_{\mathrm{state}}=32\times64\times128=262{,}144.
\label{eq:app-state-size-mamba}
\end{equation}

All recurrent architectures compared at 400M therefore have the same number of
scalar entries in their main recurrent state at each layer. At larger scales,
we preserve each method's per-head state shape and increase only the number of
heads, maintaining the state-size match across methods.

\Cref{tab:model_configurations} summarizes the resulting model configurations.
Public checkpoints retain their released architectures, tokenizers, and
context configurations and are reported separately in the result tables.

\begin{table}[tbp]
    \centering
    \caption{Configurations of the language models used in our pretraining comparisons. Public checkpoints are marked with $^{\dagger}$.}
    \label{tab:model_configurations}
    \cypatablefont
    \setlength{\tabcolsep}{12pt}
    \begin{tabular}{lccc}
        \toprule
        \textbf{Method} & \textbf{Params (M)} & \textbf{State Size} & \textbf{Configuration} \\
        \midrule
        \multicolumn{4}{l}{\textit{400M parameters with $L=24$ and $d=1024$}} \\
        \midrule
        Transformer & 374 & --      & $H=16,\ d_k=d_v=64$ \\
        SWA         & 374 & 262,144 & $H=4,\ d_k=d_v=256,\ m=128$ \\
        GSA         & 386 & 262,144 & $H=4,\ d_k=d_v=256,\ m=128$ \\
        Raven       & 412 & 262,144 & $H=4,\ d_k=d_v=256,\ m=128$ \\
        BLA         & 387 & 262,144 & $H=4,\ d_k=d_v=256,\ m^\ast=128$ \\
        Mamba-2     & 375 & 262,144 & $H=32,\ d_{\mathrm{head}}=64,\ d_{\mathrm{state}}=128$ \\
        Mamba-3     & 378 & 262,144 & $H=32,\ d_{\mathrm{head}}=64,\ d_{\mathrm{state}}=128$ \\
        GDN         & 400 & 262,144 & $H=4,\ d_k=d_v=256$ \\
        KDA         & 400 & 262,144 & $H=4,\ d_k=d_v=256$ \\
        CyFA        & 389 & 262,144 & $H=4,\ d_k=d_v=256,\ m^\ast=128$ \\
        \midrule
        \multicolumn{4}{l}{\textit{800M parameters with $L=24$ and $d=1536$}} \\
        \midrule
        Transformer  & 778 & -- & $H=24, d_k=d_v=64$ \\
        GDN         & 835 & 393,216 & $H=6,\ d_k=d_v=256$ \\
        KDA         & 816 & 393,216 & $H=6,\ d_k=d_v=256$ \\
        CyFA        & 800 & 393,216 & $H=6,\ d_k=d_v=256,\ m^\ast=128$ \\
        \midrule
        \multicolumn{4}{l}{\textit{1.4B parameters with $L=24$ and $d=2048$}} \\
        \midrule
        Transformer$^{\dagger}$ & 1,364 & --        & $H=32,\ d_k=d_v=64$ \\
        RetNet$^{\dagger}$      & 1,352 & 1,048,576 & $H=8,\ d_k=256,\ d_v=512$ \\
        GLA$^{\dagger}$         & 1,366 & 524,288   & $H=4,\ d_k=256,\ d_v=512$ \\
        GSA$^{\dagger}$         & 1,377 & 262,144   & $H=4,\ d_k=d_v=512,\ m=64$ \\
        KDA         & 1,416 & 524,288 & $H=8,\ d_k=d_v=256$ \\
        CyFA        & 1,394 & 524,288 & $H=8,\ d_k=d_v=256,\ m^\ast=128$ \\
        \bottomrule
    \end{tabular}
\end{table}

\subsection{Pretraining Configuration}

All models are implemented with the \texttt{flash-linear-attention} library
\citep{yang2024fla} and pretrained on SlimPajama-627B using the Mistral
tokenizer. Initial pretraining uses a context length of 2,048 tokens.

We optimize all models with AdamW using a peak learning rate of
$3\times10^{-4}$ and a weight decay of $0.01$. After 1,024 warmup steps, the
learning rate follows a cosine schedule to $3\times10^{-5}$. We clip the
gradient norm at $1.0$.

All runs use eight NVIDIA RTX Pro 6000 GPUs. \Cref{tab:pretraining_configuration} reports the training-token budget, effective batch size, gradient accumulation, and wall-clock time for each model scale.

\begin{table}[tbp]
    \centering
    \caption{Pretraining configurations by model scale. Per-device batch size is the number of sequences processed on each GPU before gradient accumulation.}
    \label{tab:pretraining_configuration}
    \cypatablefont
    \begin{tabular}{@{}lcccccc@{}}
        \toprule
        \textbf{Scale} & \textbf{Training Tokens} & \textbf{Context} & \makecell{\textbf{Global Batch}\\\textbf{(Tokens)}} &
        \makecell{\textbf{Per-Device}\\\textbf{Batch}} & \makecell{\textbf{Gradient}\\\textbf{Accumulation}} & \textbf{Wall Time} \\
        \midrule
        400M & 15B  & 2,048 & 0.5M & 32 & 1 & $\sim$10 hours \\
        800M & 30B  & 2,048 & 0.5M & 16 & 2 & $\sim$1 day \\
        1.4B & 100B & 2,048 & 1M   & 16 & 4 & $\sim$1 week \\
        \bottomrule
    \end{tabular}
\end{table}

\paragraph{Long-context continuation.}
For the 800M and 1.4B evaluations in \cref{sec:long_context_ability}, we resume
the pretrained checkpoints together with their optimizer states and train for
another 1,024 steps at a context length of 8,192. To preserve the number of
tokens in each optimizer step, we use a per-device batch size of 4. The 800M
run uses two gradient-accumulation steps, retaining a global batch of 0.5M
tokens, while the 1.4B run uses four steps, retaining a global batch of 1M
tokens. The learning rate remains fixed at $3\times10^{-5}$, which is the
terminal value of the preceding cosine schedule. All remaining settings are
unchanged from initial pretraining.

\section{Additional Analysis}
\label{app:additional_analysis}

\subsection{Gradient Propagation Through the Learned Clock}
\label{app:clock_gradient_analysis}

We analyze the instability observed when $\operatorname{sg}$ is removed from
\cref{eq:clock_gradient_stabilization} using paired backward probes on four
2,048-token training sequences from the 400M CyFA checkpoint. The intervention
restores only the derivative from $\delta_t$ to $\boldsymbol x_t$, leaving the
loss and all forward activations unchanged. We use KDA as a reference because
it has data-dependent recurrent control but does not use a shared cumulative
clock.

Averaged across the four sequences, CyFA with $\operatorname{sg}$ has a global
gradient norm of $5.653$, compared with $2.309$ for KDA. The difference is
concentrated in the clock projection, whose gradient norm is $5.052$ and
accounts for $78.3\%$ of the squared global norm. Excluding this projection
reduces the CyFA norm to $2.447$, only $1.06\times$ the KDA norm. The larger
global norm therefore originates primarily from the clock projection rather
than from uniformly larger gradients throughout the model.

This concentration follows from the cumulative clock. Its forward and backward
relations are
\begin{equation}
\lambda_t=\sum_{s\leq t}\delta_s,
\qquad
\frac{\partial\mathcal L}{\partial\delta_s}
=
\sum_{t\geq s}\frac{\partial\mathcal L}{\partial\lambda_t}.
\label{eq:app-clock-gradient-accumulation}
\end{equation}
Each clock increment consequently receives gradient contributions from all
later positions through the temporal write vectors of the key and value states
and through the clock-dependent slot readout. In the layers with the largest
clock gradients, the reverse cumulative sum amplifies the local
$\lambda$-gradient norm by $3.4$--$9.6\times$. Alignment among the resulting
per-token parameter gradients contributes another $7.4$--$11.8\times$. The
three clock-dependent paths are individually large and frequently oppose one
another, leaving a cancellation-sensitive residual in the shared
low-dimensional clock projection.

Removing $\operatorname{sg}$ passes this gradient accumulated from later
positions into the residual stream. The global norm rises to $12.006$ and
remains $8.682$ after excluding the clock projection, reaching $3.76\times$ the
KDA norm. The additional gradient spreads into the value write, output
projection, MLP, and earlier layers. Its effect also grows with sequence
length. Without $\operatorname{sg}$, the non-clock gradient relative to KDA
increases from $1.14\times$ at 256 tokens to $4.58\times$ at 2,048 tokens. With
$\operatorname{sg}$, it remains within $1.05$--$1.08\times$ over the same
range.

The stop-gradient operation therefore preserves the cumulative gradient that
trains the clock projection while preventing that length-dependent gradient
from propagating into the backbone.

\subsection{Input Dependence of the Learned Clock}
\label{app:clock_input_dependence}

We probe the final 400M CyFA checkpoint using the configuration in
\cref{tab:model_configurations}: 24 CyFA layers with four heads per layer,
$d=1024$, $d_k=d_v=256$, $m=127$ active slots, and $m^\ast=128$ storage rows.
We collect $\delta_t$ from 64 consecutive 2,048-token WikiText-2 sequences,
totaling 131,072 tokens and 12,582,912 head--token observations. A further 16
sequences containing 32,768 tokens are reserved for the held-out interventions.

We classify the post-sigmoid clock increments using their empirical quantiles.
A head is classified as near one when
\begin{equation}
p_{05}(\delta_t)>0.95,
\label{eq:app-near-one-clock-head}
\end{equation}
and as input dependent when
\begin{equation}
p_{95}(\delta_t)-p_{05}(\delta_t)\geq0.1.
\label{eq:app-input-dependent-clock-head}
\end{equation}
All 96 heads fall into one of these two regimes. Among them, 56 heads are
concentrated near one, with mean $0.9961$ and standard deviation $0.0065$. The
remaining 40 heads are input dependent, with mean $0.5118$ and standard
deviation $0.1627$.

Excluding the first layer, where the clock receives only the token embedding,
current-token identity explains $49.4\%$ of the variation in the
input-dependent heads. Contextual variation among occurrences of the same
token explains the remaining $50.6\%$. This decomposition is consistent with
the same-token examples in \cref{tab:clock_increment_cases}.

We then intervene on the 40 input-dependent heads while leaving the rest of the
model unchanged. The baseline held-out NLL is $2.6105$. Replacing each head's
clock increments with its mean raises NLL by $0.0450$, showing that the model
uses their variation. Cyclically shifting the original increments by 257
tokens preserves their values and marginal distribution while breaking their
alignment with the corresponding inputs. This intervention increases NLL by
$0.1238$. Setting $\delta_t=1$ throughout produces the largest increase,
$0.1381$. Every intervention degrades all 16 held-out sequences.

The larger effect of shifting the increments than replacing them with per-head
means indicates that CyFA uses their alignment with the corresponding contexts
in addition to their marginal variation.

\begin{center}
    \begin{minipage}{0.62\linewidth}
    \centering
    \captionof{table}{Held-out NLL under clock interventions in the 400M CyFA model. $\Delta\mathrm{NLL}$ is measured relative to the original clock increments.}
    \label{tab:clock_interventions}
    \cypatablefont
    \setlength{\tabcolsep}{8pt}
    \begin{tabular}{lcc}
        \toprule
        \textbf{Clock Setting} & \textbf{NLL} & $\boldsymbol{\Delta}$\textbf{NLL} \\
        \midrule
        Original $\delta_t$ & 2.6105 & -- \\
        Per-head mean & 2.6555 & $+0.0450$ \\
        Shifted by 257 tokens & 2.7344 & $+0.1238$ \\
        $\delta_t=1$ & 2.7486 & $+0.1381$ \\
        \bottomrule
    \end{tabular}
    \end{minipage}
\end{center}

\section{Additional Experimental Results}
\label{app:additional_experimental_results}

\begin{figure}[htbp]
    \centering
    \includegraphics[width=\linewidth]{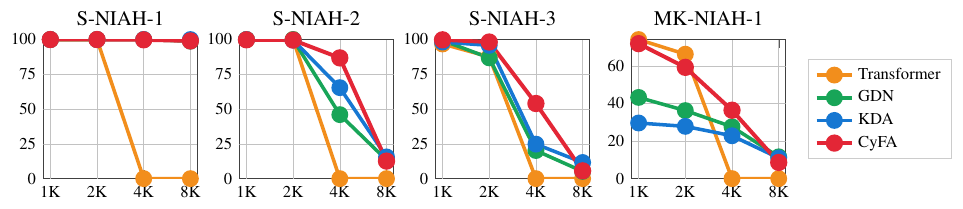}
    \caption{RULER accuracy of the 800M models before long-context continuation.}
    \label{fig:ruler_before_continuation}
\end{figure}

\begin{table}[htbp]
    \centering
    \caption{LongBench results before long-context continuation, with inputs truncated to 8K tokens.}
    \label{tab:longbench_before_8k}
    \cypatablefont
    \setlength{\tabcolsep}{2.8pt}
    \renewcommand{\arraystretch}{1.05}
    \resizebox{\linewidth}{!}{%
    \begin{tabular}{l|l|cc|ccc|ccc|ccc|ccc|c}
        \toprule
        \multirow[c]{2}{*}[-0.4ex]{\textbf{Scale}} & \multirow[c]{2}{*}[-0.4ex]{\textbf{Model}} & \multicolumn{2}{c|}{Code} & \multicolumn{3}{c|}{Summarization} & \multicolumn{3}{c|}{SingleQA} & \multicolumn{3}{c|}{MultiQA} & \multicolumn{3}{c|}{Few-Shot} & \multirow{2}{*}{\textbf{Avg.}} \\
        \cmidrule{3-16}
        & & \textbf{LCC} & \textbf{RBP} & \textbf{GvR} & \textbf{QMS} & \textbf{MNs} & \textbf{NQA} & \textbf{QQA} & \textbf{MQA} & \textbf{HQA} & \textbf{2WM} & \textbf{MSQ} & \textbf{TRE} & \textbf{TQA} & \textbf{SAM} & \\
        \midrule
        \multirow{4}{*}{800M} & Transformer & 22.08 & 6.40 & 0.74 & 0.74 & 0.67 & 0.00 & 0.73 & 1.67 & 0.32 & 0.21 & 0.00 & 1.50 & 0.78 & 0.15 & 2.57 \\
        & GDN & \underline{43.96} & 37.53 & \textbf{5.82} & 10.64 & \underline{2.26} & \textbf{4.34} & 1.81 & \textbf{13.07} & \textbf{4.88} & \textbf{10.03} & 2.12 & \underline{27.00} & 25.29 & \textbf{9.11} & 14.13 \\
        & KDA & 42.16 & \underline{39.23} & \underline{3.87} & \underline{11.78} & 2.09 & \underline{3.43} & \underline{5.79} & 12.51 & 4.08 & 7.69 & \underline{2.62} & \textbf{40.50} & \underline{25.65} & \underline{5.82} & \underline{14.80} \\
        & CyFA & \textbf{47.31} & \textbf{43.92} & 3.15 & \textbf{15.99} & \textbf{5.64} & 2.71 & \textbf{5.81} & \underline{13.06} & \underline{4.76} & \underline{9.25} & \textbf{2.94} & 23.00 & \textbf{38.40} & 4.59 & \textbf{15.75} \\
        \midrule
        \multirow{2}{*}{1.4B} & KDA & \underline{46.12} & \underline{42.70} & \underline{4.44} & \textbf{14.05} & \underline{8.68} & \underline{2.13} & \underline{0.17} & \underline{7.72} & \underline{6.27} & \underline{7.88} & \underline{2.90} & \textbf{53.50} & \underline{41.11} & \textbf{28.97} & \underline{19.05} \\
        & CyFA & \textbf{52.65} & \textbf{45.07} & \textbf{12.58} & \underline{13.91} & \textbf{13.52} & \textbf{2.79} & \textbf{5.26} & \textbf{14.11} & \textbf{7.54} & \textbf{12.46} & \textbf{3.44} & \underline{37.50} & \textbf{48.67} & \underline{17.13} & \textbf{20.47} \\
        \bottomrule
    \end{tabular}}
\end{table}

\end{document}